# Pinned and Still Unstable: Within-Judge Verdict Variance and the Noise Floor of LLM-as-Judge Leaderboards

**Authors:** Krishna Chytanya Ayyagari[1]
[1]Google

## Abstract

Modern LLM evaluation depends on the assumption that pinning the judge to a specific model snapshot and setting the decoding temperature to zero produces reproducible verdicts. We show that this assumption fails as a property of how LLM-as-Judge is *operationalized via cloud serving infrastructure* — demonstrated here on one major enterprise cloud platform (named in §3.2) — rather than of a single model family. Our experiments quantify the benchmark-level *effects* of this non-determinism; we do not isolate its root causes, and we treat the system-level mechanisms documented for temperature-zero cloud inference in prior work (floating-point non-associativity in batched matmul, MoE routing dependent on co-located requests, hardware-revision drift) as the *hypothesized* sources rather than ones we establish here. Whether other providers' caching, batching, and routing behave the same is left open. Across four frontier judges all served via a single major enterprise cloud platform (§3.2) — `gemini-2.5-pro` as the principal reference judge, `claude-haiku-4-5` for cross-family confirmation, and `claude-sonnet-4-5` / `claude-sonnet-4-6` as a matched-snapshot pair for cross-snapshot drift analysis — we quantify how much within-judge variance survives pinning and temperature-zero decoding on standard leaderboards (MT-Bench, AlpacaEval 2, Arena-Hard), and what that variance does and does not do to the resulting rankings.

Pilot studies on 50 and 80 prompts from Arena-Hard already establish four claims: (1) on the two pilot judge/pair configurations, frontier judges exhibit per-item verdict flip rates in the **3–8% range** across 10 independent re-runs, with **30–38% conditional flip rates** on the close-call items where leaderboards are most sensitive (the full sweep below reveals a wider *per-judge* range, from 0.13% to 9.7%, once more judges and candidates are included); (2) under a position-swap protocol with identical content on reversed A/B assignment, flip rates are **statistically identical** (e.g. 2.86% vs 2.89% on one judge) and the aggregate winner is invariant to A/B order — so the run-to-run variance operates independently of any consistent position preference on this disparate-tier pair (a residual position component on near-tie pairs is not fully excluded; §8); (3) judge stability differs markedly across judges — some waver less often but with higher conditional intensity, others waver more often — establishing that judge variance is not a single number but a per-judge profile (whether the cross-family gap is intrinsic to the model family or partly a serving-stack effect we cannot resolve on a single platform; §8); (4) verdict-format compliance varies enough between judge families that structured-output enforcement (`response_schema` for Gemini, forced `tool_use` for Anthropic) is required for reliable measurement.

The full Stage 2 sweep (principal judge `gemini-2.5-pro`, 10 candidates, 10 independent re-runs) confirms the pilot effect at full scale and sharpens what it does to leaderboards. Per-item verdict flip rates are **4.9%** on Arena-Hard and **5.3%** on AlpacaEval 2, rising to a **~40% conditional flip rate** on the items where the judge ever wavers; on MT-Bench — where scores compress against the ceiling (mean 9.58 ± 0.04) — pairwise leadership between two models flips on **5.4%** of turn-1 comparisons across re-runs (6.5% if the less-well-specified, context-free turn-2 items are pooled in; §6.2). This within-judge noise is genuine at the item level and at the margins, yet it largely averages out of the aggregate: for the principal judge every top-K set is **identical across all ten re-runs** and no candidate's win/loss against the baseline changes sign (0% top-K instability, 0% aggregate winner flip) — though adjacent middle ranks can still swap across re-runs (e.g. the Llama-4-Maverick vs. Llama-3.3-70B pair, §6.5). What the noise removes is not the ranking but its *pre-* *cision* — under a hierarchical bootstrap over prompts and runs, **4–6 of every 9 adjacent leaderboard positions are statistically indistinguishable** for the principal judge (2–7 of 9 across all four judges; §6.5) — a precision limit a decomposition attributes to benchmark prompt sampling rather than judge variance (§6.4). The headline is therefore not that a single judge inverts its own leaderboard on re-run — it does not — but that the leaderboard's noise floor leaves much of its fine-grained ordering statistically unresolved. Extending to a second family and a matched-snapshot pair (`claude-haiku-4-5`, `claude-sonnet-4-5`, `claude-sonnet-4-6`), the effect behaves as a *per-judge property*: within-judge stability spans a 40× range (Haiku flips 0.13% of Arena-Hard verdicts; Sonnet-4-5 flips 5–10%), and the noisier Sonnet judges reshuffle even their own top-3 or top-5 sets on roughly half of re-run pairs (§6.5c). Across judges, leaderboards agree on the coarse ordering but diverge on the fine-grained ranking — Kendall's $\tau$ of 0.42–0.64 between Gemini and the Sonnet judges on Arena-Hard (≈18–29% of model pairs ranked oppositely; other cross-family pairs agree more; the Arena-Hard values are sensitive to candidate answers cut off by the generation cap, §8), and MT-Bench absolute scores that differ by ∼2 points depending on which judge is used. Re-judging 15 expected head-to-head orderings (from vendor model tiers, model generations, public leaderboard standings, and two expectations of our own; §3.5), **12 survive every re-run of the principal judge but only 8 survive every judge that evaluated them** — so **a judge from a second family overturns one-third of the orderings a single judge reproduces**; 2 are contradicted by every judge (one robust to length control, one length-confounded; §6.5).

We argue current LLM-as-Judge leaderboards report unhedged point estimates that do not convey the noise floor of the underlying measurement. We propose a minimal reporting protocol — R≥5 re-runs per judge, published flip-rate confidence intervals, adjacency-separability disclosure — adoptable at marginal cost relative to the candidate-training budgets the leaderboards are used to compare.

## 1. Introduction

The release of MT-Bench (Zheng et al., 2023) made LLM-as-Judge the dominant evaluation methodology for open-ended generation. Two years later, many frontier-model releases, post-training method papers, and alignment techniques are justified in part by LLM-judged leaderboard scores. Practitioners treat these scores as stable: a model that ranks third on Arena-Hard is "the third-best model on Arena-Hard," and a 1.2-point improvement is reported as a result.

This treatment depends on a load-bearing assumption: **that the judge is a stable measurement instrument**. The assumption is so embedded in the methodology that papers rarely state it explicitly, and almost none report confidence intervals over judge variance. The two standard hedges — pinning the judge to a versioned snapshot and setting decoding temperature to zero — are widely believed to make the assumption hold.

We do not argue that LLM-as-Judge is broken or that current leaderboards should be discarded. We argue that the standard mitigation **is incomplete in a measurable way**: even with both hedges in place, repeated runs of the same pinned, T=0 judge on identical inputs produce residual variance large enough to flip the verdict on close-call items — exactly the items where competing models are similar in true quality and where leaderboard headline rankings are decided. Most leaderboard rows are stable; the close-call rows that drive interpretation are not. This is a problem about how leaderboards **report** their numbers, not about whether the underlying methodology has value. The fix is straightforward and cheap: report uncertainty intervals — over judge re-runs *and* over the benchmark's finite prompt sample — instead of unhedged point estimates. (As §6.4 shows, the two sources are distinct: run-to-run judge variance drives item-level and head-to-head instability, while the aggregate adjacency noise floor is dominated by prompt sampling; an honest leaderboard must hedge for both.)

Our contributions:

1. We introduce **per-item flip rate**, **conditional intensity**, **pairwise outcome flip rate**, **top-K instability**, and **adjacency separability** (formally defined in §3.4) as the metrics that matter for leaderboard consumers, and argue they are more decision-relevant than the score-correlation metrics (Spearman $\rho$, Cohen's $\kappa$) dominant in the existing judge-reliability literature.
2. Across three benchmarks (MT-Bench, AlpacaEval 2, Arena-Hard), four judges from two model families served on one platform, and $M = 10$ candidate models (§3.3), we show that pinned, temperature-zero judges are non-deterministic, and that the magnitude is a **per-judge property spanning ~40×** (`claude-haiku-4-5` flips 0.13% of Arena-Hard verdicts; `claude-sonnet-4-5` flips 5–10%) — so judge variance must be reported as a profile (waver fraction *and* conditional intensity, reported separately), not a single scalar. The reference judge's aggregate ranking is stable on re-run; separately, between a fifth and three-quarters (2–7 of 9) of adjacent leaderboard positions are statistically indistinguishable — a benchmark prompt-sampling limit that a deterministic judge would share (§6.4), which we distinguish from the genuine judge effects — while the noisier judges do reshuffle even their own top-K sets across identical re-runs (§6.5c).
3. We show leaderboards **disagree across judges**: Gemini-vs-Sonnet Kendall's $\tau$ of 0.42–0.64 on Arena-Hard (≈18–29% of model pairs ordered oppositely; these Arena-Hard values are sensitive to truncated candidate answers, §8), while agreeing on the coarse ordering. Re-evaluating $C = 15$ expected head-to-head orderings (§3.5) the way the leaderboards themselves rank, **8 of the 15 survive every judge that evaluated them** while 12 survive the principal judge alone, so a second judge overturns one-third of the orderings a single judge reproduces; the 7 failures decompose into cross-judge disagreement (3), within-judge instability (2), and unanimous contradiction (2, one of them length-confounded; §6.5).
4. We propose a minimal reporting protocol: publish a flip-rate confidence interval from $R \geq 5$ judge re-runs, a per-judge stability profile, an adjacency-separability disclosure, and — given cross-judge disagreement — results under more than one judge (§7.3).

## 2. Related Work

**LLM-as-Judge benchmarks.** Zheng et al. (2023) introduced MT-Bench and Chatbot Arena, and reported >80% agreement between GPT-4 judges and human pairwise preferences — establishing the empirical case for the LLM-as-Judge methodology that subsequent work built on. Dubois et al. (2023) introduced AlpacaFarm and the original AlpacaEval; Dubois et al. (2024) introduced the length-controlled (LC) variant of AlpacaEval 2 to debias against verbose outputs (we use the raw, non-LC win-rate; §3.1). Li et al. (2024) introduced Arena-Hard, distilling Chatbot Arena's live user prompts into a 500-prompt benchmark with a fixed reference baseline. Zeng et al. (2024) introduced LLMBar to measure judges on adversarial instruction-following cases.

**Judge biases.** A substantial literature documents systematic biases in single LLM judge runs. Wang et al. (2023) demonstrated position bias and proposed swap-based mitigations. Saito et al. (2023) and Dubois et al. (2024) quantified verbosity bias. Panickssery et al. (2024) showed that LLM judges recognize and prefer their own outputs. Chen et al. (2024) compare human and LLM judgment biases head-to-head. Stureborg et al. (2024) report within-run inconsistency in scalar-rating judges. Thakur et al. (2024) systematically catalog prompt sensitivity and adversarial vulnerabilities of judge models. Liu et al. (2024) compare scalar and pairwise judging protocols. Chen and Goldfarb-Tarrant (2025) show that LLM safety evaluators are not robust to superficial input artifacts, flipping verdicts without a change in substance. Closest in spirit to our protocol, Schroeder and Wood-Doughty (2024) quantify LLM-judge reliability with a psychometric coefficient (McDonald's $\omega$), show it degrades with temperature, and argue against single-shot evaluation — but they measure reliability as a scalar and do not trace it through to leaderboard *rankings* or head-to-head claims. These works characterize the *direction* of judge errors, or reliability as an aggregate; to our knowledge, none quantifies how much verdict variance survives the standard mitigations (pinning + temperature zero) at the item level, or what that residual variance does to fine-grained *rankings* — which is our focus.

**Temporal drift.** Chen et al. (2023) showed that ChatGPT's behavior on generation tasks shifts substantially over months. Our work extends the drift question to *judging* behavior, and — more importantly — measures within-snapshot variance under nominally deterministic decoding, which is invisible to drift studies that compare snapshot means.

**Determinism failures.** It is now widely understood that temperature-zero LLM inference is not bitwise reproducible, due to floating-point non-associativity in batched matrix multiplications, mixture-of-experts routing dependent on co-located requests, and provider-side load-balancing across hardware (Chann, 2023; subsequent provider documentation). Atil et al. (2024) quantified the *downstream* effect for general tasks, finding accuracy swings of up to 15% across repeated runs of five nominally-deterministic LLMs on eight tasks. Our work specializes this to the LLM-as-Judge setting and carries it one step further — from output non-determinism to its effect on *pairwise verdicts and leaderboard rankings* — which, to our knowledge, has not been measured.

**Leaderboard brittleness and finite-sample noise.** A parallel line of work notes that benchmark rankings are noisy simply because benchmarks are finite samples of prompts. Bowyer et al. (2025) show that central-limit-theorem confidence intervals dramatically underestimate uncertainty on small benchmarks; Gonzalez et al. (2025) find single-run leaderboards brittle, with most rank slices flipping relative to a multi-run majority. Our adjacent-position noise floor (§6.4) is consistent with — and we credit to — this prompt-sampling source rather than claiming it as novel. On the theoretical side, Wang (2026) derives fundamental limits on how well any finite audit can certify a model in the rare-error regime — a complementary bound on what evaluation can establish in principle, where our contribution is an empirical measurement of one such limit (the leaderboard noise floor and the judge's share of it) on today's leaderboards. Our distinct contribution is to isolate the *judge* dimension layered on top of prompt-sampling noise: within-judge run-to-run verdict variance (§6.2), its disagreement *across* judges (§6.5), and its effect on expected head-to-head orderings (§6.5).

**Gap.** No prior work — including the task-level non-determinism of Atil et al. (2024) or the finite-sample brittleness of Bowyer et al. (2025) and Gonzalez et al. (2025) — measures the rate at which leaderboard pairwise outcomes *flip* between independent runs of an identically-configured judge, and no prior work translates judge-level variance into a confidence interval over published rankings. The existing reliability literature reports per-item agreement statistics (Cohen's $\kappa$, Spearman correlations) which compress flips at the top of the leaderboard with stable agreement at the bottom into a single average. We argue this compression is exactly what hides the failure mode that matters to consumers of these benchmarks.

## 3. Methodology

### 3.1 Benchmarks

We use three benchmarks chosen for their prevalence in published post-training work:

- **MT-Bench** (Zheng et al., 2023): 80 two-turn prompts across 8 categories. We score each turn as an independent single-turn item (160 items total), each rated 1–10 by a single judge. Each item is judged in isolation: the judge sees only that turn's question and answer, with **no turn-1 context supplied for turn-2 items**. This isolates single-turn judgment variance but means turn-2 items that presuppose turn-1 (e.g. "revise your previous answer") are judged without that context; we treat this as a scope choice and revisit it in §8.

- **AlpacaEval 2** (Dubois et al., 2024): 805 single-turn prompts, pairwise-judged against the fixed reference `gpt-4-1106-preview`. We use the **raw** pairwise win-rate and do **not** apply the length-controlled (LC) GLM adjustment, because our object of study is judge-verdict variance, not the calibrated absolute win-rate; "AlpacaEval 2" in this paper therefore denotes the un-length-controlled win-rate. The LC adjustment reweights the *aggregate* win-rate but changes no individual verdict, so it leaves every per-verdict variance metric (§6.2) untouched; it can bear only on aggregate-derived AlpacaEval quantities (top-K churn, cross-judge $\tau$, and the *replication* survival; §6.5). For the replication survival we separately verify that the outcomes are robust to length control except for #12, which we flag as length-confounded, and #5, whose cross-judge disagreement disappears under one of our two length adjustments.
- **Arena-Hard v0.1** (Li et al., 2024): 500 single-turn prompts sourced from Chatbot Arena, pairwise-judged against the fixed reference baseline `gpt-4-0314` (the upstream Arena-Hard anchor).

### 3.2 Judges

Four judge configurations, all served via a single major enterprise cloud platform — in our experiments, Google Cloud's Gemini Enterprise Agent Platform (formerly Vertex AI), which we refer to as "the platform" throughout — at `temperature=0`:

- `gemini-2.5-pro` — frontier Gemini judge (principal)
- `claude-haiku-4-5` — Claude Haiku, served via the platform's Anthropic offering (global endpoint), used for cross-family confirmation at minimal API cost
- `claude-sonnet-4-5` — Claude Sonnet snapshot A, served via the platform's Anthropic offering, paired with the next judge for cross-snapshot drift analysis
- `claude-sonnet-4-6` — Claude Sonnet snapshot B, served via the platform's Anthropic offering, the matched-tier next-snapshot judge that enables a clean cross-snapshot drift comparison without confounding scale and snapshot

In the Stage 2 sweep, judges are invoked by platform model alias rather than by a dated snapshot identifier; the model version reported by the serving endpoint was constant within each judge throughout the sweep. "Pinned" in this paper means fixed to that version (§8).

For each judge we run the full benchmark with **R=10 independent re-runs** per item. The two pilot studies and the Stage 2 sweep use successive verdict-extraction configurations:

- **Tier 0 sniff test (§4.1)** uses the canonical Arena-Hard prompt template with the verdict requested as `[[A]] | [[B]] | [[C]]`, parsed from the judge's free-form text.
- **Tier 1 study (§4.2)** uses an adapted prompt that requests the verdict as one of `A`, `B`, or `tie`, but **also parsed from free-form text** — structured-output enforcement was not yet in place. As reported in §4.3, this configuration produced ~24% verdict-format parse failures on `gemini-2.5-pro`, which motivated the move to structured output for Stage 2.
- **Stage 2 sweep (§5)** enforces structured output at the API layer: `response_schema` (with the verdict field constrained to `A | B | tie`) for Gemini judges, and forced `tool_use` with a `submit_pairwise_verdict` tool for Anthropic judges. Text-parsing is retained as a belt-and-suspenders fallback. The underlying judging task (impartial comparison of two responses to the same prompt) is unchanged across all three configurations; only the verdict-extraction mechanism differs.

We choose a single-platform design deliberately. First, it removes cross-provider infrastructure variance from the judge stage: all four judges are served from the same cloud control plane with the same authentication, retry, and rate-limiting characteristics. This is conservative — any flips we observe cannot be attributed to inter-provider differences in batching or load-balancing. Second, the judge mechanisms we hypothesize as variance sources (floating-point non-associativity in batched matrix multiplications, MoE routing dependent on co-located requests, hardware-revision drift) are not provider-specific, so a finding established under one platform carries the standard methodological burden of needing to be re-checked elsewhere — but the burden is light, because the underlying mechanism is universal. We stress that this choice is **not a critique of the platform's engineering**: the non-determinism we measure is consistent with properties of batched, load-balanced GPU inference documented in prior work (§7.2), and the platform is in fact what makes the study feasible — it is the one environment in which we could serve Gemini, Claude, and open-weight judges under a single control plane, holding the serving stack fixed across model families. The protocol itself is platform-agnostic: we used one platform for a controlled, apples-to-apples comparison, and replicating the study on other providers' stacks (e.g. OpenAI, Azure OpenAI, AWS Bedrock, or each vendor's native API) is straightforward — we did not do so here and flag it as natural future work (§8). We address the implications in §7.

We had originally planned to include cross-snapshot drift comparisons within the Gemini Pro family (`gemini-2.5-pro` vs an older Gemini Pro snapshot); that older snapshot was deprecated from the platform during this study. Cross-snapshot drift within the Gemini Pro family is therefore left to future work pending availability of matched snapshots. The Claude Sonnet 4.5 / 4.6 pair (J3, J4) preserves the cross-snapshot drift hypothesis in a different family with the same experimental structure.

### 3.3 Candidate Models

We evaluate $M = 10$ candidate models spanning four model families and a range of capability tiers, all served on the same platform as the judges:

- **Gemini family (3):** `gemini-2.5-pro`, `gemini-2.5-flash`, `gemini-2.5-flash-lite` (the lightest model serves as a de facto weak-model floor)
- **Claude family (3):** `claude-haiku-4-5`, `claude-opus-4-1`, `claude-opus-4-5`
- **Open-weights via the platform's Model Garden managed endpoints (MaaS) (4):** `meta/llama-4-maverick-17b-128e-instruct-maas`, `meta/llama-3.3-70b-instruct-maas`, `qwen/qwen3-235b-a22b-instruct-2507-maas`, `qwen/qwen3-next-80b-a3b-instruct-maas` (all served pay-per-token; we deliberately use only managed/MaaS endpoints, no self-deployed models)

The pool deliberately spans four model vendors (Google, Anthropic, and two independent open-weights vendors — Meta and Alibaba) so that cross-family agreement and family-clustering effects (§5.3) can be measured against models belonging to neither judge family. The platform model identifiers (for the Gemini and Claude models, aliases rather than dated snapshots, as for the judges; §3.2) and the per-candidate generation configuration (T=0, `max_output_tokens=2048`, region) are listed in `config.py` (Appendix A). The 2,048-token cap cuts off some of the longer Arena-Hard answers (§8). The exact set was selected from the models served on the platform at the time of the sweep, and earlier-generation snapshots that were retired from the platform during the study were replaced with their nearest served successors.

### 3.4 Metrics

**Notation.** Let $M$ be the number of candidate models, $R$ the number of independent judge re-runs per item ($R = 10$ throughout this paper), and let $i$ index the benchmark items. For the pairwise benchmarks an "item" is a (prompt, candidate) comparison *cell* — so the 5,000 Arena-Hard items are 500 prompts × 10 candidates and the 8,050 AlpacaEval items are 805 × 10 — while for MT-Bench an item is a (prompt-turn, model) score; the per-item metrics below, and the cell counts in §6.2, are over these items.

For **pairwise benchmarks (Arena-Hard, AlpacaEval 2)** we define the verdict $V_i^{(r)}(m, m') \in \{A, B, \text{tie}\}$ returned by run $r$ comparing model $m$ (slot B) against model $m'$ (slot A) — in every Stage 2 comparison the reference answer is shown first, in slot A, and the candidate second, a fixed rather than randomized order (§8). For AlpacaEval 2, $m'$ is the fixed reference model `gpt-4-1106-preview` (for Arena-Hard, $m'$ is the fixed reference `gpt-4-0314`; §3.1); the same definitions reduce to a fixed-baseline comparison. We aggregate the **raw** win-rate $\bar{W}$, not the length-controlled GLM adjustment (§3.1). From $V$ we derive a **win indicator** that explicitly handles ternary outcomes:

$$W_i^{(r)}(m, m') = \begin{cases} 1 & V_i^{(r)} = B \\ 0 & V_i^{(r)} = A \\ 0.5 & V_i^{(r)} = \text{tie} \end{cases}$$

Ties count as half-credit for win-rate *aggregation* (the convention used by Arena-Hard and AlpacaEval upstream). For the *flip* metric below, however, we count any change in the verdict as a flip: since $W \in \{0, 0.5, 1\}$, a $B \to \textbf{tie}$ transition $(1 \to 0.5)$ satisfies $W^{(r)} \neq W^{(r')}$ and is counted as a flip, identical to $B \to A$. Ties are therefore never silently absorbed, but the flip metric measures verdict *instability* (did the verdict change), not the magnitude of the change.

For **scalar benchmarks (MT-Bench)**, let $S_i^{(r)}(m) \in \{1, \ldots, 10\}$ be the integer score assigned to model $m$ on item $i$ in run $r$. We define the scalar leadership indicator $L_i^{(r)}(m, m') = \textbf{sign}(S_i^{(r)}(m) - S_i^{(r)}(m'))$, which encodes which model is preferred $(+1, -1)$ or scored equal $(0)$. Because each model's scores come from independent single-model judge calls, matching the run index $r$ across $m$ and $m'$ is one valid pairing of i.i.d. samples and yields $R$ leadership signs per item; a full $R \times R$ cross-pairing would use more comparisons and give a lower-variance estimate of the leadership-flip rate, at the cost of dependence among the pairs. We use matched-index pairing for simplicity.

**Metrics.**

- **Per-item flip rate (pairwise).** For models $m, m'$, the per-item disagreement rate — for a given item $i$, the fraction of the $\binom{R_i}{2}$ run pairs on which the win indicator differs — then averaged over benchmark items:

$$\text{FlipRate}_i(m, m') = \frac{1}{\binom{R_i}{2}} \sum_{r<r'} \mathbb{1}\left[W_i^{(r)}(m, m') \neq W_i^{(r')}(m, m')\right]$$

  reported as the mean over benchmark items. We use $R_i$ in place of $R$ to make explicit that the denominator adapts to the number of *successfully parsed* runs per item — under the Tier 1 text-format configuration, parse failures left some items with $R_i < R$, and the calculation in §4.2 uses each item's own $R_i$ (the flip rate is undefined for $R_i < 2$, so such items are excluded).
- **Per-item flip rate (scalar / MT-Bench).** Same definition with $W$ replaced by the leadership indicator $L \in \{-1, 0, +1\}$: the fraction of items where the *leadership* between $m$ and $m'$ changes between two runs. As in the pairwise case this is a binary change count, so a shift from a clear lead to a tie $(+1 \to 0)$ counts the same as a full reversal $(+1 \to -1)$ — it records whether leadership moved, not by how much.
- **Waver fraction and conditional intensity.** The **waver fraction** is the share of items whose per-item flip rate is nonzero — items where the verdict changed on at least one of the $\binom{R_i}{2}$ run pairs: **waver** $= \frac{1}{N_{\text{items}}} \sum_i \mathbb{1}[\text{FlipRate}_i > 0]$ (we write $N_{\text{items}}$ for the item count to avoid collision with the rerun count $R$). The **conditional intensity** is the mean per-item flip rate restricted to those wavering items: $\mathbf{cond} = \mathbf{mean}\{\mathbf{FlipRate}_i : \mathbf{FlipRate}_i > 0\}$. The mean per-item flip rate factors as $\mathbf{waver} \times \mathbf{cond}$; we report the two separately as the judge's **stability profile** (§7.5). The waver fraction increases with $R$ (a larger $R$ exposes more items as ever-wavering), so profiles are only comparable at equal $R$.

- **Pairwise outcome flip rate.** For each model pair, the fraction of run pairs $(r, r')$ for which the *aggregate* winner (computed by averaging $W$ across items) differs. Computed as the mean over all $\binom{R}{2}$ run pairs.
- **Top-K instability.** Probability that the *unordered set* of top-K models changes between two independent runs (internal reshuffling within the top-K does not count as a change), where ranking is computed by aggregating $W$ across items per run (for the scalar benchmark, the per-run mean score $\bar{S}^{(r)}(m) = \frac{1}{N_{\text{items}}} \sum_i S_i^{(r)}(m)$). For $K = M$ this is trivially zero (the set is all $M$ models), so with $M = 10$ we report $K \in \{1, 3, 5\}$. When two models tie exactly at the K/K+1 boundary on a run, that run's top-K set is ambiguous; where this occurs we report the range over tie-breaks (§6.5c).
- **Kendall's $\tau$.** Rank correlation between two leaderboards, computed as Kendall's $\tau$-b (which corrects for ties). We report it both *within-judge* — across the $\binom{R}{2}$ run pairs, summarizing re-run agreement (§6.3) — and *cross-judge*, between two judges' aggregate rankings (§6.5d).
- **Adjacent-position separability.** For each adjacent pair on the leaderboard, the probability that their order is preserved under a **paired hierarchical bootstrap**: we resample benchmark prompts with replacement (outer level, applied *jointly* to both models so the comparison stays paired and their positive cross-prompt correlation is preserved) and, within each resampled prompt, draw a **single** judge run (inner level) — modeling the one-evaluation-per-prompt design of a standard leaderboard, rather than averaging the $R$ runs, which would suppress single-run judge variance by a factor of $R$ (§6.4) — then recompute each model's aggregate score. The *order-preservation probability* is the fraction of 20,000 resamples in which the higher-ranked model still outscores the lower-ranked one; we call a pair **indistinguishable** when this falls below 0.95 — i.e. the higher-ranked model fails a *one-sided* 95% test that it is truly ahead (equivalently, the 90% two-sided bootstrap interval of the paired score difference includes zero). Under the stricter two-sided-95% convention (order-preservation threshold 0.975) the counts are identical or at most one pair larger on every judge, so the noise floor is not an artifact of the threshold. Because the bootstrap resamples both prompts and runs, this interval reflects prompt-sampling *and* judge run-to-run uncertainty together; §6.4 decomposes the two.

### 3.5 Replication Study

We assemble $C = 15$ expected head-to-head orderings among models in our candidate pool (so their outputs are available), each evaluated on one of the three benchmarks. Each has a stated basis (Appendix C): vendor model tiering (7 orderings), model generation (2), public leaderboard standing at the time of access (4), or our own expectation (2, both AlpacaEval). These are expectations, not reported results: the tiering and generation orderings reflect how vendors position their own models rather than results on Arena-Hard, AlpacaEval 2, or MT-Bench. For each ordering, we re-evaluate the comparison the way the leaderboard ranks (win-rate vs. the fixed baseline for Arena-Hard/AlpacaEval, mean score for MT-Bench) under our R-run protocol and report whether the expected ordering survives, distinguishing genuine run-to-run instability from stable judge-disagreement. The same procedure extends to published claims from surveyed papers, including those whose models fall outside the pool, by supplying their published outputs.

### 3.6 What We Are Measuring (and What We Are Not)

We measure variance attributable to the judge alone. We hold candidate model outputs fixed (generated once, temperature=0, cached) so that all observed variance is isolated to the judge stage. We did not apply cache-busting to the judge calls. The nonzero run-to-run variance we observe — including `claude-haiku-4-5`'s 0.13%, which would be exactly zero under a 100% cache-hit rate — rules out *deterministic (full) caching*. It does not rule out *partial* caching (e.g. occasional hits from regional load-balancing or periodic eviction), which would only *suppress* the measured flip rates; our rates are therefore a **lower bound**. A prompt-level cache-buster (a semantically-neutral nonce) would remove this ambiguity and is a cheap addition for future runs. Holding candidate outputs fixed is conservative: real benchmark workflows regenerate candidate outputs, which adds further variance on top of what we report.

## 4. Pilot Studies

We report two pilot studies, conducted at a total cost of approximately $33 in API spend, that establish the central effect, rule out the most plausible alternative explanation (position bias), and surface a methodological finding (verdict-format compliance varies enough between judge families to warrant structured-output enforcement). The pilot runner code and per-run verdict data are part of the artifact bundle described in Appendix A.

### 4.1 Pilot 1 — Single-Judge Sniff Test (Tier 0)

**Setup.** 50 prompts from Arena-Hard v0.1 (deterministic uid-sorted subset). Two candidate models (`gemini-2.5-flash` as Assistant A, `gemini-2.5-flash-lite` as Assistant B), both generated once at T=0 with responses cached for the duration of the experiment. Single judge: `gemini-2.5-pro` on the platform, `temperature=0`, `thinking_budget=128` (the minimum that the model accepts), `max_output_tokens=4096`. The judge was invoked **R=10** independent times per prompt with identical inputs and verdict format `[[A]] | [[B]] | [[C]]`.

**Result.**

| Metric | Value |
|---|---|
| Items judged | 50 |
| Independent re-runs per item | 10 |
| **Mean per-item verdict flip rate** | **6.4%** |
| % items unanimous across all 10 runs | 80.0% |
| Items non-unanimous (close-call subset) | 10 |
| **Mean flip rate on non-unanimous items** | **32.2%** |
| Aggregate winner (per re-run) | Stable across all 10 |

**Interpretation.** A pinned, T=0 frontier judge given identical inputs returned a different verdict on 6.4% of items across 10 independent re-runs. The judge wavered on 20% of items; within that non-unanimous subset it contradicted itself on roughly one in three re-run pairs (a conditional intensity of 32.2%). The aggregate winner was stable — as the full sweep later confirms it remains (§6.3) — but the per-item disagreement rate is high enough to matter on the close-call comparisons that decide leaderboard margins.

### 4.2 Pilot 2 — Cross-Family Confirmation and Position-Swap (Tier 1)

**Setup.** 80 prompts from Arena-Hard v0.1, two pairs (Pair 1: A=`gemini-2.5-flash`, B=`gemini-2.5-flash-lite`; Pair 2: same content with A and B reversed), two judges (`gemini-2.5-pro`, regional endpoint; `claude-haiku-4-5@20251001`, global endpoint via the platform's Anthropic offering), R=10 re-runs per (judge × pair × prompt). All other settings as in Pilot 1, **except the verdict format**, which moves from Pilot 1's Tier 0 (`[[A]] | [[B]] | [[C]]`) to Tier 1 (`A | B | tie`, per §3.2).

**Result.**

| Judge | Pair | Items | Per-item flip rate | Conditional flip rate (non-unanimous items) | % unanimous |
|---|---|---|---|---|---|
| Judge $J_1$ | A=Flash, B=Lite | 79 | 6.2% | 35.1% | 82.3% |
| Judge $J_1$ | A=Lite, B=Flash | 80 | 7.6% | 32.2% | 76.3% |
| Judge $J_2$ | A=Flash, B=Lite | 80 | 2.86% | 38.1% | 92.5% |
| Judge $J_2$ | A=Lite, B=Flash | 80 | 2.89% | 38.5% | 92.5% |

We use the compact labels $J_1$ and $J_2$ for the two judges in this table; both are identified in the setup above, and their full configurations are given in Appendix B. $J_1$ Pair 1 reports 79 items rather than 80 because one item failed to parse on every re-run under the Tier 1 text-format configuration and was therefore excluded (see §4.3).

**Three findings emerge:**

1. **Cross-family generalization.** Both judges, drawn from two different model families and served under different platform endpoint configurations, exhibit non-trivial per-item flip rates. The effect is therefore a property of the LLM-as-Judge methodology under pinned + T=0 inference, not an artifact of any one model family or endpoint configuration (both judges were served on the same platform; §8).
2. **Position bias is ruled out on this pair.** Pair 2 is Pair 1 with A and B reversed; the judge sees identical content with the only change being position. Per-judge flip rates are statistically indistinguishable across the two pairs (e.g. $J_2$: 2.86% vs 2.89%; difference well within sampling noise at N=80). More directly, **the aggregate winner identity is invariant under position swap**: in both pairs and for both judges, the same candidate model wins all 10 re-runs regardless of A/B assignment. A pure-position-bias hypothesis predicts the aggregate winner should depend on which model is in position B (or A); observationally, it does not. Per-judge variance is therefore real and acts on top of any consistent preference, not in place of it. We caution that this pair is disparate-tier (Flash vs. Flash-Lite): a consistent quality preference could mask a residual position effect on *near-tie* pairs, which we did not swap at scale (§8).
3. **Judge stability is a profile, not a single number.** $J_1$ wavers on a larger fraction of items (~21% non-unanimous; 17.7–23.7% across the two A/B orders) at moderate intensity (~33% conditional); $J_2$ wavers on a smaller fraction (~7.5% non-unanimous) at higher intensity (~38% conditional). A reader who collapsed these to a single "judge stability" scalar would lose load-bearing information.

### 4.3 Methodological finding: structured output is required for reliable measurement

In the Tier 1 setup with text-format verdict parsing (verdict requested as `A | B | tie`, per §3.2), one judge's verdict-format compliance was substantially lower than the other's (~24% parse failures vs ~0%). Investigation showed that when the judge's full response (including any internal-reasoning prelude) exceeded `max_output_tokens=4096`, the verdict token at the tail was truncated and the response could not be parsed. We note that a pairwise judgment explanation that runs to >4,000 tokens is anomalous — typical pairwise rationales on this prompt template are a few hundred tokens — and we attribute the long tail

of failed responses to a degenerate generation pathology under temperature-zero decoding (a well-documented failure mode in which the same low-perplexity continuation is repeated until a stopping criterion is hit; recent work proposes decoding penalties specifically to suppress it, e.g. Ginart et al., 2025), not to a natural response-length distribution that happens to overlap the token cap. In other words: the 24% number reflects a *generation* bug exposed by text-format parsing, not the judge's natural verbosity. (This is a known, model-agnostic autoregressive failure mode, and here it is induced by the deliberately minimal `thinking_budget=128` we set on this judge (§4.1) — it is *not* representative of the model's default configuration, and a standard reasoning budget would likely remove it; we flag this in §8.) Structured-output enforcement (schema-constrained decoding, e.g. Willard & Louf, 2023) is an established mitigation for such degenerate generation and is not itself a novel finding of this paper; we adopt it as a measurement prerequisite. One caveat: format restrictions can themselves alter a model's behavior on reasoning-style tasks (Tam et al., 2024), so a schema-enforced judge could in principle differ from a free-form one — but the pilots (§4.2), which used free-form text parsing, exhibit the same qualitative variance, so the effect we measure is not an artifact of schema enforcement. Schema-constrained decoding keeps the explanation-then-verdict order (so the judge's chain-of-thought is preserved — the verdict is *not* forced to be emitted first), but it constrains the decoder to produce a syntactically complete, parseable `verdict` field rather than a free-form tail token that can be truncated, and empirically it eliminated the parse failures (≈0% vs ~24%). We are careful about the mechanism: a repetition loop *inside* a JSON string value would still run to the token cap and fail to parse, so the near-total elimination suggests the schema-constrained decoding path also steers generation away from the degenerate attractor, not merely that it wraps a still-looping model in valid syntax. Either way the effect is empirical, and the reasoning-then-verdict order is preserved (the verdict is not forced first). The flip-rate numbers reported for that judge in §4.2 are therefore computed across the items where verdicts did parse successfully (per-call parse success rate ≈ 76%, with most items retaining ≥9 of 10 valid verdicts); we did not find these failures concentrated on any identifiable prompt subset, and Stage 2's schema-constrained output all but removed them (at least 99.8% of verdicts are valid for every judge × benchmark; §6.5), so the headline Stage-2 results rest on essentially complete data and only these pilot rates rely on partially-parsed items. To remove this confound for the Stage 2 sweep, we adopt structured output at the API layer:

- **For `vertex_gemini` judges:** `response_mime_type="application/json"` with a `response_schema` that requires `{"explanation": str, "verdict": "A"|"B"|"tie"}`.
- **For `vertex_anthropic` judges:** forced `tool_use` with a `submit_pairwise_verdict` tool whose `input_schema` requires the same fields.

We have validated the structured-output path on each judge during pilot smoke checks; per-judge verdict format compliance under structured output is verified by a smoke check before each Stage 2 sweep and re-checked after every re-run, with a hard abort if a re-run's parse-error rate exceeds 1% (0.5% after the first re-run). **The use of structured output here is a methodological prerequisite for reliable measurement of judge variance, not a model-quality finding** — text-format verdicts under unconstrained decoding will produce parse errors that confound the analysis on any judge whose response length distribution overlaps the token cap. We recommend any future LLM-as-Judge benchmark adopt schema-enforced verdict extraction as standard practice.

## 5. Stage 2 Experimental Design

The four experiments below extend the pilot studies to publication scale; each either (a) generalizes a pilot finding across a broader experimental matrix, or (b) measures a quantity the pilots could not address at N=50–80. All four were run; results are reported in §6, and each experiment below points to the corre-sponding results subsection.

### 5.1 Experiment 1: Within-Snapshot Variance at Scale

**Setup.** Pin each of the four judges to its target model version (§3.2), `temperature=0`. Run the full evaluation on **all three benchmarks** (Arena-Hard 500 prompts, AlpacaEval 2 805 prompts, MT-Bench 80 prompts × 2 turns = 160 single-turn items) at $\boldsymbol{R = 10}$ independent re-runs each, on the fixed 10-candidate matrix from §3.3. For each benchmark and each judge we measure per-item flip rate and pairwise outcome flip rate at population scale (the scalar variant defined in §3.4 applies to MT-Bench). This experiment confirms that the Pilot 1 / Pilot 2 effect generalizes from the 80-prompt Arena-Hard subset to the full benchmark and across all four judges, and extends it to two benchmarks the pilots did not touch.

**Why it matters.** Pilot 2 established the effect at N=80 on one pair structure; the full leaderboard claim requires it at benchmark scale across the full candidate matrix.

**Result (§6.2, §6.5a).** Confirmed: the effect generalizes from the 80-prompt pilot to all three benchmarks and all four judges, with per-item flip rate a per-judge property spanning ∼40× (0.13% for Haiku to 5–10% for Sonnet-4-5).

### 5.2 Experiment 2: Cross-Snapshot Drift Within the Claude Sonnet Family

**Setup.** Same as 5.1 but compare aggregate rankings produced by `claude-sonnet-4-5` and `claude-sonnet-4-6` — two adjacent snapshots within the same model tier. Decompose total variance into within-snapshot (5.1) and across-snapshot components.

**Hypothesis and scope.** `sonnet-4-5` and `sonnet-4-6` are two *intentionally distinct* released versions, not the same endpoint sampled at two times, so a ranking difference between them is expected and *confirms* the value of pinning (an unpinned judge that silently advanced from 4-5 to 4-6 would shift rankings). Our claim is narrower and about *magnitude*: we ask how a deliberate one-step version change compares to the within-snapshot re-run variance each snapshot already exhibits under pinning. If the two are comparable, then even a pinned judge's residual re-run noise is on the same scale as a version bump — so pinning removes the version axis but leaves a non-trivial noise axis, and longitudinal score comparisons across versions must account for both.

**Result (§6.5d).** The two snapshots produce rankings that differ (Arena-Hard Kendall's $\tau$-b = 0.78, ≈5 of 45 pairs reordered) and, more to the point, have markedly different *within-snapshot* stability profiles (Arena-Hard per-item flip 5.16% vs. 3.57%; MT-Bench turn-1 leadership flip 12.15% vs. 14.71%) — so pinning to either version fixes the version but not the per-snapshot re-run noise, and the stability profile itself is version-dependent.

### 5.3 Experiment 3: Cross-Family Judge Substitution

**Setup.** Using the four-judge ensemble (`gemini-2.5-pro`, `claude-haiku-4-5`, `claude-sonnet-4-5`, `claude-sonnet-4-6`), report Kendall's $\tau$ between aggregate rankings produced by each judge, plus overlap of top-K sets. Pilot 2 established that within-snapshot variance is non-trivial across two judge families with **distinct stability profiles per judge** (different waver fractions and different conditional intensities, see §4.2 and §7.5, Discussion); this experiment characterizes the cross-judge ranking agreement at the leaderboard level, including the within-Sonnet-snapshot agreement (J3 vs J4) which provides a tighter bound than the cross-family comparison.

**Result (§6.5d).** Cross-judge Kendall's $\tau$ ranges 0.42–0.87 on Arena-Hard (0.87–0.96 on AlpacaEval): judges agree on the coarse ordering but Gemini and the Sonnet judges reorder ≈18–29% of Arena-Hard model pairs, the largest disagreement being cross-family. A possible judge self-preference case surfaces (only Gemini ranks Gemini-2.5-Pro above Qwen3-235B on Arena-Hard). Both observations are sensitive to truncated candidate answers (§8).

### 5.4 Experiment 4: Replication of Expected Head-to-Head Orderings

**Setup.** For each of C=15 expected head-to-head orderings (§3.5), re-evaluate the comparison the way the leaderboard ranks — win-rate vs. the fixed baseline (Arena-Hard/AlpacaEval) or mean score (MT-Bench) — across R=10 re-runs of each judge, reusing the main-sweep verdicts at no extra cost. Report: (a) did the ordering reproduce on an arbitrary single run (we report run $r = 1$; the runs are i.i.d., so there is no chronological "first")? (b) what fraction of the R=10 re-runs reproduce it? (c) does it survive *every* judge, or do judges disagree on the winner?

**Reporting protocol.** Population-level statistics ("X of 15 claims did not survive re-judging") and the per-claim failure table are reported in §6.5 (Table 1), framed as robustness of the expected ordering to re-runs and judge choice — not as a correctness verdict on any source.

**Result (§6.5).** 8 of the 15 claims survive every judge that evaluated them (12 under the principal judge alone); the 7 failures are cross-judge disagreement (3), within-judge instability (2), and unanimous contradiction (2).

## 6. Results

### 6.1 Setup

Stage 2 ran the principal judge `gemini-2.5-pro` over all three benchmarks at full scale: 500 Arena-Hard prompts and 805 AlpacaEval 2 prompts — each of 10 candidate models judged against the fixed baseline, giving 5,000 and 8,050 prompt×candidate comparison cells respectively — and 80 MT-Bench prompts with each of the two turns scored separately (160 single-turn items) for each of 10 models. Every cell was re-judged across **R=10 independent re-runs** under a single pinned snapshot at temperature 0, with structured output enforced (`response_schema` for Gemini, forced `tool_use` for Anthropic). We then repeated the identical protocol with three further judges — `claude-haiku-4-5` (cross-family), and `claude-sonnet-4-5` / `claude-sonnet-4-6` (a matched-snapshot pair). Haiku and Sonnet-4-5 completed all three benchmarks; Sonnet-4-6 completed Arena-Hard and MT-Bench, and its AlpacaEval run was terminated early and is excluded throughout. §6.2–6.4 use `gemini-2.5-pro` as the reference judge; §6.5 reports the per-judge comparison and the cross-judge results.

### 6.2 Within-judge verdict variance is real and substantial

| Pairwise benchmark | Cells (R=10 runs) | Per-item flip | Waver fraction | Conditional intensity |
|---|---|---|---|---|
| Arena-Hard | 5,000 | 4.94% | 12.5% | **39.6%** |
| AlpacaEval 2 | 8,050 | 5.29% | 13.5% | **39.1%** |

(MT-Bench uses absolute scoring rather than pairwise verdicts, so it does not share these columns; its numbers — mean score 9.58 ± 0.04, per-item leadership-flip 5.39% turn-1 / 6.46% pooled — are given in the text below.)

Under a pinned snapshot at temperature 0, the mean per-item flip rate — the probability that two independent re-runs disagree, averaged over comparison cells — is **4.94%** on Arena-Hard and **5.29%** on AlpacaEval. This is not the fraction of *unstable* cells: most cells never waver, and this mean averages the many stable cells (flip rate 0) with the minority that do. Separating the two, the fraction of cells that ever flip (the *waver fraction*) is 12.5% on Arena-Hard and 13.5% on AlpacaEval, and *restricted to those wavering cells* the **conditional flip intensity is 39.6% and 39.1%** respectively. On the items where the judge wavers — which we take as a *proxy* for close calls, while noting that prompt ambiguity or the judge's own stochastic rationale generation (candidate outputs are held fixed, §3.6, so the wavering is not from candidate regeneration) can also drive it — roughly two of every five run-pairs disagree. On MT-Bench, where absolute scores compress against the ceiling (mean 9.58 ± 0.04 across models and runs), the per-item leadership between two models flips on **5.39%** of turn-1 comparisons for the principal judge, averaged over the 45 model pairs. We take the **turn-1-only** rate as the primary MT-Bench figure because turn-2 items are judged without their turn-1 context (§3.1) and are therefore underspecified: they are about twice as unstable on per-item score dispersion (score-std 0.11 vs. 0.21 for the principal judge; waver fraction 11.4% vs. 17.7%; the same gap holds for every judge) and ~40% more unstable on leadership flip (5.39% turn-1 vs. 7.54% turn-2), so *pooling* the two turns inflates the figure to 6.46% and conflates genuine judge variance with the underspecification of context-free turn-2 prompts. Turn-1, turn-2, and pooled leadership-flip rates for all four judges are given in §6.5a; the pooled rate runs ~20–40% above turn-1 throughout. We return to this design choice in §8.

### 6.3 …but the aggregate ranking does not invert on re-run

The within-judge noise does not propagate to the aggregate leaderboard. For the principal judge, across all 45 run-pairs (the $\binom{10}{2}$ pairs of the 10 re-runs):

- **Aggregate pairwise winner flip: 0.0%** on both Arena-Hard and AlpacaEval. No candidate's pooled win/loss against the baseline changes sign across re-runs.
- **Top-K instability: 0.0%** for K ∈ {1, 3, 5}. The top-1, top-3, and top-5 *sets* are identical across every re-run. (We omit K=10, which is trivially stable when the pool is exactly M=10 models.)

The item-level noise of §6.2 largely averages out at benchmark scale: pooling ~500–805 per-prompt verdicts stabilizes the aggregate even though individual verdicts are unstable. This is the paper's central honest qualification — **the reference judge does not invert its own leaderboard on re-run.** Two things are stable across all four judges: every candidate's win/loss *sign* against the baseline (0% aggregate winner flip), and the top-1 model. What is *not* perfectly stable is the **full ordering** of adjacent, near-tied models — even for Gemini, the Llama-4-Maverick vs. Llama-3.3-70B pair swaps relative order on 4 of 10 re-runs (§6.5, Claim #8), a mid-leaderboard swap that leaves the top-K sets and win/loss signs untouched — and the Sonnet judges, and Haiku at a near-tied 5/6 boundary, additionally reshuffle their top-3 or top-5 sets (§6.5c). Consistent with this, the within-judge Kendall's $\tau$-b between any two re-run leaderboards on Arena-Hard averages 0.96 (Gemini), 0.98 (Haiku), 0.90 (Sonnet-4-5), and 0.92 (Sonnet-4-6) — high, but dipping as low as 0.69 for the noisiest judge's worst run-pair, i.e. the same mid-leaderboard churn.

### 6.4 The noise floor: adjacent positions are statistically indistinguishable

What the variance changes is the *precision* the leaderboard implies. We model a standard single-evaluation leaderboard — one judge run per prompt — and put a confidence interval on it with the **paired** hierarchical bootstrap of §3.4: resample prompts jointly across models (outer), then draw exactly **one** of the $R$ judge runs per resampled prompt (inner). For each adjacent pair we ask whether the higher-ranked model still outscores the lower one; a pair is **indistinguishable** when order-preservation falls below 0.95 — the higher model fails a one-sided 95% test that it is truly ahead (equivalently, the 90% two-sided interval of their *paired* score difference includes zero; the stricter two-sided-95% threshold, 0.975, changes the counts by at most one pair per judge). (This paired test is deliberately not the weaker "do the two models' marginal confidence intervals overlap" heuristic, which ignores the positive cross-prompt correlation between adjacent models and would overstate indistinguishability.) Of the 9 adjacent pairs in each 10-model leader-board, **6 of 9 are indistinguishable on Arena-Hard and 4 of 9 on AlpacaEval** for the principal judge — with one Arena-Hard pair sitting essentially on the 0.95 boundary ($p \approx 0.95$), so that count is 6–7 by a hair. Between four and six of every nine adjacent positions are inside the instrument's noise floor.

To separate the two noise sources — the benchmark's finite prompt sample and the judge's run-to-run variance — we re-run the bootstrap in three modes: **prompt-only** (each model's runs collapsed to their per-prompt mean, i.e. a deterministic judge, then prompts resampled), **run-only** (prompts fixed, one run drawn per prompt — the single-run judge variance in isolation), and **both** (the single-eval bootstrap above).

| Benchmark | Judge | prompt-only | run-only | both |
|---|---|---|---|---|
| Arena-Hard | Gemini | 6/9 | 3/9 | 6/9 |
| Arena-Hard | Haiku | 7/9 | 3/9 | 7/9 |
| Arena-Hard | Sonnet-4-5 | 5/9 | 4/9 | 6/9 |
| Arena-Hard | Sonnet-4-6 | 7/9 | 4/9 | 7/9 |
| AlpacaEval | Gemini | 4/9 | 2/9 | 4/9 |
| AlpacaEval | Haiku | 2/9 | 0/9 | 2/9 |
| AlpacaEval | Sonnet-4-5 | 4/9 | 2/9 | 4/9 |

Two things follow. First, single-run judge variance is **not** negligible in isolation: on its own (prompts fixed, one run per prompt) it makes 3–4 of 9 Arena-Hard pairs and up to 2 of 9 AlpacaEval pairs indistinguishable. (An earlier version of this decomposition averaged all $R$ runs per prompt in the inner bootstrap, which suppresses single-run variance $\sim R$-fold and understated this contribution to ≤3/9; drawing a single run per prompt, as a real leaderboard does, is the correct model and is what we now report.) Second, and despite that, the judge's *marginal* contribution once prompt sampling is also present is small: `both − prompt-only` is **0 in six of the seven cells and +1 in one** (Sonnet-4-5 on Arena-Hard). The pairs the judge blurs are largely a subset of those prompt sampling already blurs, so the combined single-eval noise floor is **dominated by the finite prompt sample**, with judge run-to-run variance adding at most one further indistinguishable pair; a perfectly deterministic judge would leave nearly the same large share (5–7 of 9) of adjacent Arena-Hard pairs unresolved. We report the noise floor because it bounds what a leaderboard's fine-grained ordering can mean, and we attribute it *primarily to prompt sampling*, not to the judge. The judge-specific effects of this paper live elsewhere: item-level flip rates (§6.2), head-to-head replication instability (§6.5, Claim #8), cross-judge disagreement (§6.5d), and the noisier judges' top-K churn (§6.5c). (Reproduced by `noise_floor_decomposition.py`.)

One caveat on interpretation: the *absolute* fraction of indistinguishable adjacent pairs depends on the capability density of this specific 10-model pool — a pool spread across widely separated tiers would show fewer indistinguishable pairs, a denser pool more. The count should therefore be read as a property of *this leaderboard* (these models on this benchmark), not a universal benchmark constant; what generalizes is the qualitative finding that a substantial share of adjacent, closely-matched models are unresolved once uncertainty is reported.

### 6.5 Cross-judge results: variance is a per-judge property

Repeating the protocol across four judges (three families, one matched-snapshot pair) turns the single-judge picture of §6.2–6.4 into three cross-judge findings.

**(a) The effect replicates across families, but its magnitude is judge-specific — a ~40× range.** Every judge exhibits the pinned-and-still-unstable effect, yet how much it wavers is a per-judge property, not a constant.

| Judge | Arena flip | Arena cond. | AlpacaEval flip | AlpacaEval cond. | MT-Bench flip (turn-1) | MT-Bench mean |
|---|---|---|---|---|---|---|
| `gemini-2.5-pro` | 4.94% | 39.6% | 5.29% | 39.1% | 5.39% | 9.58 |
| `claude-haiku-4-5` | **0.13%** | 34.0% | **0.24%** | 35.1% | **0.74%** | 7.90 |
| `claude-sonnet-4-5` | 5.16% | 37.8% | **9.71%** | 40.3% | 12.15% | 8.33 |
| `claude-sonnet-4-6` * | 3.57% | 39.8% | — | — | 14.71% | 7.64 |

*Arena-Hard + MT-Bench only (AlpacaEval run terminated early). The MT-Bench column reports the **turn-1-only** leadership-flip rate (§6.2, the cleaner estimate); the pooled two-turn rates — inflated by the context-free turn-2 items — are 6.46% / 0.96% / 13.73% / 16.18% respectively, and turn-2-only rates are 7.54% / 1.18% / 15.31% / 17.65%.

Haiku is nearly deterministic (0.13–0.24% of verdicts flip); Sonnet-4-5 is the noisiest (up to 9.71% on AlpacaEval) — a ~40× spread across judges served on the same infrastructure. The *conditional* intensity, by contrast, is near-uniform (34–40%), and it must be interpreted carefully. For $R = 10$ independent runs, an item on which one outcome (say a win for the higher-ranked model) is returned on $a$ of the runs has flip rate $a(R-a)/\binom{R}{2} = a(10-a)/45$ across run-pairs, symmetric in $a$ and $R-a$; a wavering item has $a \in \{1, \dots, 9\}$. (This binary form assumes each wavering item splits between two outcomes; a genuine ternary A/B/tie split with counts $a, b, c$ has $ab + ac + bc$ disagreeing run-pairs, but empirically wavering is near-binary, so the binary reference is a close approximation.) If this vote count were *broadly spread* — $a$ roughly uniform over $\{1, \dots, 9\}$ — the mean conditional flip rate would be $(R+1)/(3(R-1)) \approx 40.7\%$ at $R = 10$. (We parametrize by the outcome count $a \in \{1, \dots, R-1\}$, not the *minority* count, which cannot exceed $R/2 = 5$.) This is **not** what an arbitrary wavering process yields: one dominated by *barely*-split items (a single dissenting run, $a \in \{1, 9\}$) would give only $9/45 \approx 20\%$. Empirically the wavering items are broadly split rather than piled at a lone dissent — among Gemini's wavering Arena-Hard cells only **26%** have a single dissenting run ($a \in \{1, 9\}$), while **74% split more evenly and half reach a 3–7 split or closer** (Sonnet-4-5 and AlpacaEval are similar) — so the observed ~34–40% is an *empirical* fact about the close-call set (its ambiguity is broadly, near-uniformly distributed), not a trivial combinatorial constant. What makes the conditional intensity carry little *judge-discriminating* signal is that this ambiguity distribution is similar across judges, so the intensity is near-constant while the judge-specific signal lives almost entirely in the **waver fraction** — how often a judge wavers at all — which is what varies ~40×. Two reporting caveats follow: the ~40.7% reference value is **$R$-dependent** ($(R+1)/(3(R-1))$ is 50% at $R = 5$, 40.7% at $R = 10$, 36.8% at $R = 20$), so conditional intensities are comparable only at equal $R$, and they reflect the ambiguity distribution, not the judge. This refines the pilot's claim (3): judge variance is a profile, but its informative component is the waver fraction.

**(b) The noise floor holds for every judge.** Adjacent-position indistinguishability (hierarchical bootstrap, §6.4) is pervasive across judges: of 9 adjacent pairs, **6/9 (Gemini), 7/9 (Haiku), 6/9 (Sonnet-4-5), 7/9 (Sonnet-4-6) are indistinguishable on Arena-Hard**, and 4/9, 2/9, 4/9 respectively on AlpacaEval. Between one-fifth and three-quarters of every judge's leaderboard is inside its own noise floor.

**(c) The noisier judges reshuffle even their own top-K.** Gemini shows 0% top-K set instability at every reported K (1, 3, 5) — its aggregate ordering is stable on re-run (§6.3). But **Sonnet-4-5's top-3 set changes on 51–62% of re-run pairs on Arena-Hard** (top-5 on 38%), and Sonnet-4-6's top-5 on 53% (its top-3 set is stable), *while aggregate winner-flip stays 0%*. The within-judge "leaderboard reshuffles on re-run" claim — which we retract for Gemini — therefore holds for the noisier judges, as rank churn among near-tied middle positions rather than as candidates crossing the win/loss line. (Haiku is a special case worth clarifying: despite a 0.13% item-flip rate it shows top-5 instability of 36–56%. Because top-K instability is measured across re-runs on a *fixed* prompt set, a perfectly deterministic judge would score exactly 0 here — so this *is* genuine run-to-run volatility, not the prompt-sampling noise floor of §6.4. Its magnitude is amplified by two candidates near-tied at the 5/6 boundary — exactly tied on three of the ten runs — where even Haiku's rare flips suffice to swap the top-5 set.) **Tie handling.** When two candidates tie exactly at the K/K+1 boundary on a run, that run's top-K set is ambiguous (§3.4), so we report the range over tie-breaks: Haiku's top-5 instability is 36–56% (three tied runs) and Sonnet-4-5's top-3 is 51–62% (one tied run). Figure 2 plots the lower values (36% and 51%). No other reported top-K value involves a boundary tie.

**(d) Across judges, leaderboards agree coarsely but diverge in the middle.** Cross-judge Kendall's $\tau$ (n=10 candidates; $\approx(1-\tau)/2 \times 45$ model pairs ranked in opposite order):

| Judge pair | Arena-Hard $\tau$ (p) | AlpacaEval $\tau$ (p) |
|---|---|---|
| Gemini – Haiku | 0.87 (p<0.001) | 0.87 (p<0.001) |
| Gemini – Sonnet-4-5 | 0.64 (p=0.009) | 0.96 (p<0.001) |
| Gemini – Sonnet-4-6* | 0.42 (p=0.11) | — |
| Haiku – Sonnet-4-5 | 0.78 (p=0.001) | 0.91 (p<0.001) |
| Haiku – Sonnet-4-6* | 0.56 (p=0.03) | — |
| Sonnet-4-5 – Sonnet-4-6* | 0.78 (p=0.001) | — |

$\tau$ is Kendall's $\tau$-b (ties-corrected; n=10 candidates); p-values are shown, and only the Gemini–Sonnet-4-6 pair (p=0.11) is not statistically significant.

*Arena-Hard + MT-Bench only; Sonnet-4-6's AlpacaEval run did not complete, so AlpacaEval $\tau$ values involving it are omitted.

On Arena-Hard, cross-family judges reorder a meaningful minority of the leaderboard: at $\tau$=0.64 (Gemini vs Sonnet-4-5) ≈8 of 45 model pairs are ranked oppositely, and at $\tau$=0.42 (Gemini vs Sonnet-4-6, not significant) ≈13 of 45. AlpacaEval agreement is higher ($\tau$ 0.87–0.96). Every within-family pair and most cross-family pairs are significantly positively correlated — so judges agree on the coarse ordering — but the disagreement concentrates in the middle of the leaderboard, the same noise-floor region where within-judge adjacency fails (§6.4). This is cross-judge *disagreement*, not wholesale inversion: swapping the judge reorders roughly a fifth to a quarter of Arena-Hard model pairs. These Arena-Hard values are sensitive to candidate answers cut off by the generation cap, which judges may score differently (§8).

**(e) On MT-Bench, judges disagree on the scale itself.** Mean absolute scores span **7.64 (Sonnet-4-6) to 9.58 (Gemini)** — a ~2-point gap on a 10-point scale — and the turn-1 leadership-flip rate varies ~20× (0.74% Haiku to 14.71% Sonnet-4-6; 0.96%–16.18% pooling both turns). An absolute MT-Bench number is therefore not comparable across studies unless the judge is named.

**Replication survey.** We re-evaluate C=15 expected head-to-head orderings (§3.5; the basis for each is in Appendix C) the way the leaderboards themselves rank, i.e. win-rate vs. the fixed baseline (Arena-Hard/AlpacaEval) or mean score (MT-Bench), reusing the sweep verdicts at no extra cost. Under the **prin-** **cipal judge alone, 12 of 15 survive** all 10 re-runs (#3, #8, and #12 do not); **across every judge that evaluated each claim, 8 of 15 survive** (the two AlpacaEval survivors, #7 and #11, are confirmed on the three judges with AlpacaEval data, since Sonnet-4-6 lacks that benchmark). Introducing judges from a second family therefore overturns **4 of the 12 (one-third)** that the principal judge reproduces: #2, #5, #6, and #9. (Per-claim first-run reproduction and the fraction of the 10 re-runs that reproduced each claim are recorded in `summary_from_sweep.json` (Appendix A); the within-judge failures below are exactly the claims whose reproduced fraction is under 1.)

Our *survival* criterion is deliberately strict — the claimed ordering must hold in **all 10 re-runs of all judges** — because the subject is reproducibility, not the point estimate. A strict count should not be misread as claiming every failure is an *inversion* of the expected winner: under a softer **aggregate** criterion (the majority-of-runs winner matches the claim for every judge), **10 of 15** hold. The two claims that clear the aggregate bar but fail the strict one — #2 (Sonnet-4-5 reproduces 9/10) and #8 (Gemini 6/10) — are the within-judge-instability cases: the expected winner is unchanged, but a single re-run can flip it. The remaining five (#3, #5, #6, #9, #12) fail even the aggregate criterion, because at least one judge's majority winner is the *other* model. We headline the strict count because an ordering that a re-run can silently flip is exactly the fragility this paper documents, but we flag the distinction so a 9/10 is not read as an inversion. The 7 strict failures split into three mechanisms (Table 1):

- **Cross-judge disagreement (3/15):** different judges name different winners for the same comparison — #5 (Gemini Flash vs. Flash-Lite), #6 (Claude Opus 4.5 vs. 4.1), and #9 (Gemini-2.5-Pro vs. Qwen3-235B on Arena-Hard). #9 is notable: the *only* judge that ranks Gemini-2.5-Pro above Qwen3-235B is Gemini itself — all three Claude judges rank Qwen3-235B higher, a pattern consistent with judge self-preference. It may also reflect differences in how judges score truncated candidate answers (§8); we do not separate the two explanations.
- **Within-judge instability (2/15):** the winner is agreed across judges but is not stable within one — #2 (Sonnet-4-5) and #8, the Llama-4-Maverick > Llama-3.3-70B generation ordering, which Gemini flips on 4 of 10 identical re-runs (a near-tie pair inside the §6.4 noise floor).
- **Unanimous contradiction (2/15):** every judge with AlpacaEval data (Gemini, Haiku, Sonnet-4-5) consistently ranks the *other* model above the claimed winner — for #3, Qwen3-Next-80B over Qwen3-235B; for #12, Qwen3-235B over Claude Opus 4.5. Both are stable *disagreements with the claim*, not run-to-run variance. They behave differently under length control, however (next paragraph): #3 is a genuine contradiction, while #12 is length-confounded.

**Length-control robustness (AlpacaEval).** AlpacaEval's official leaderboard ranks by the *length-controlled* (LC) win-rate, whereas we rank by the raw win-rate (§3.1), so a natural concern is that AlpacaEval replication failures are verbosity artifacts rather than judge effects. This concern does **not** touch the *item-level* metrics: the LC adjustment reweights the *aggregate* win-rate but changes no individual verdict, so per-item flip, waver fraction, and conditional intensity (§6.2, §6.5a) are length-control-invariant. It *can*, however, shift any **aggregate-derived** quantity — top-K churn, cross-judge $\tau$ (§6.5c–d), and the replication survival — since those are computed from win-rates that LC reweights; our AlpacaEval top-K and $\tau$ values are reported on the raw metric, and a length-controlled leaderboard could reorder them (a scope limitation we flag in §8). Where the concern is sharpest — the replication survival — we re-checked each AlpacaEval claim directly under two transparent length-adjusted win-rates: a **length-matched subset** (prompts where the model's output is within ±25% of the baseline's character length) and a **per-model logistic length adjustment** evaluated at zero length difference. The outcome is claim-specific. The within-family contradiction **#3** (Qwen3-Next-80B preferred over the larger Qwen3-235B) is **robust**: Next-80B wins under the raw and both length-controlled win-rates for all three judges — and, although it is the *longer* model (so a raw preference for it is exactly what verbosity would predict), it keeps winning *after* length control, which is what rules verbosity out as the explanation. In contrast **#12** is **length-confounded**: Claude Opus 4.5's outputs are ~43% shorter than Qwen3-235B's (1,471 vs. 2,577 characters), and under length control the ordering becomes judge- and method-dependent (e.g. Gemini and Sonnet-4-5 prefer Opus 4.5 on the length-matched subset), so #12 is not a clean contradiction — we count it among the failures but flag it as metric-dependent. Among the remaining AlpacaEval claims, **#8** (Llama-4-Maverick > Llama-3.3-70B) is a genuine near-tie (raw win-rates 0.396 vs. 0.393) whose ordering is expectedly fragile to both re-runs and length control — consistent with its sitting inside the noise floor — while **#7, #11** survive under raw and both length controls for all three judges with AlpacaEval data (Gemini, Haiku, Sonnet-4-5). **#5** is itself partly length-dependent: its cross-judge disagreement persists on the length-matched subset (Haiku still prefers Flash-Lite), but under the logistic adjustment Haiku prefers Flash (0.863 vs. 0.854), so all three judges agree with the claim under that variant. (Reproduced by `alpaca_length_control.py`.)

**Table 1. Claims failing to survive re-judging (8 of 15 survive; the 7 failures shown).**

| # | Claim | Benchmark | Failure mode |
|---|---|---|---|
| 5 | Gemini 2.5 Flash > Flash-Lite | AlpacaEval | cross-judge disagreement (Haiku ranks Flash-Lite higher — reproduces 0/10; Sonnet-4-5 reproduces 9/10) |
| 6 | Claude Opus 4.5 > Opus 4.1 | Arena-Hard | cross-judge disagreement (Sonnet-4-5 & Sonnet-4-6 rank Opus 4.1 higher; Haiku split 5/5; Gemini reproduces) |
| 9 | Gemini 2.5 Pro > Qwen3-235B | Arena-Hard | cross-judge disagreement (all 3 Claude judges rank Qwen3-235B higher; possibly self-preference, see text) |
| 2 | Claude Opus 4.5 > Haiku 4.5 | Arena-Hard | within-judge instability (Sonnet-4-5 flips across re-runs) |
| 8 | Llama-4-Maverick > Llama-3.3-70B | AlpacaEval | within-judge instability (Gemini flips on 4/10 re-runs) |
| 3 | Qwen3-235B > Qwen3-Next-80B | AlpacaEval | unanimous contradiction (all judges rank Next-80B higher) |
| 12 | Claude Opus 4.5 > Qwen3-235B | AlpacaEval | unanimous contradiction (all judges rank Qwen3-235B higher) |

(Sonnet-4-6 is excluded from all AlpacaEval results — its AlpacaEval run did not complete — so the AlpacaEval failures above are established on the judges with complete data.)

Position-swap robustness (identical content on reversed A/B assignment) was established on the pilot (§4.2: 2.86% vs 2.89%); the full-scale swap is an optional confirmation, not a load-bearing result.

**Data completeness.** Gemini, Haiku, and Sonnet-4-5 completed all assigned cells; Sonnet-4-6 completed Arena-Hard and MT-Bench only (its AlpacaEval run was cut short when we stopped the sweep to bound *total* spend, and is excluded, along with any AlpacaEval $\tau$ involving it). Completing that one remaining cell (a single judge × benchmark) would itself be inexpensive and would close the matched-snapshot pair. Within the completed cells, at least 99.8% of verdicts are valid for every judge × benchmark (25 and 94 of the 50,000 Arena-Hard verdicts are missing for Sonnet-4-5 and Sonnet-4-6, and 2 and 1 of the 16,000 MT-Bench scores for Gemini and Sonnet-4-6); missing verdicts are excluded from the per-item metrics.

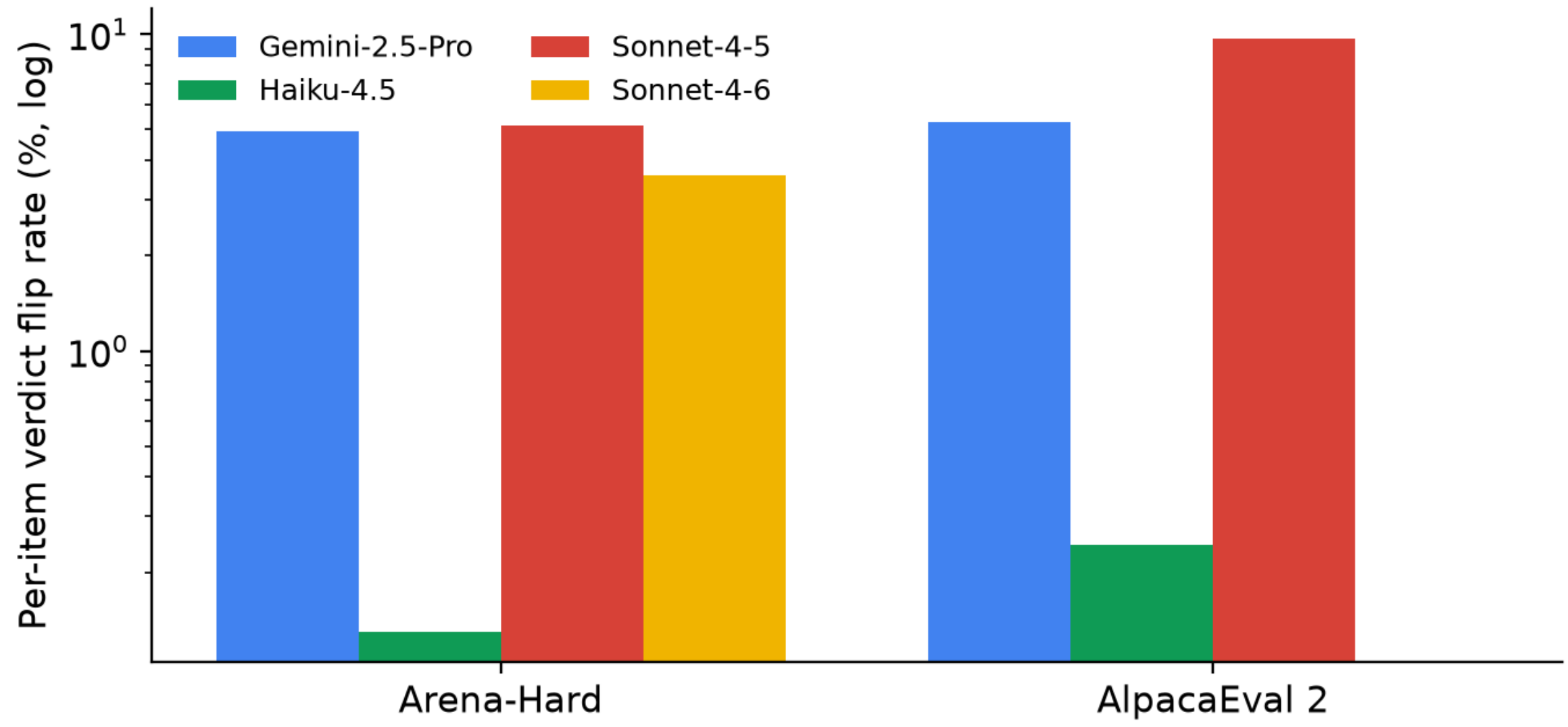

**Figure 1.** Per-item verdict flip rate by judge (log scale), per benchmark — the ∼40× cross-judge stability range (§6.5a). Haiku is near-deterministic (0.13–0.24%); the Sonnet judges are the noisiest.

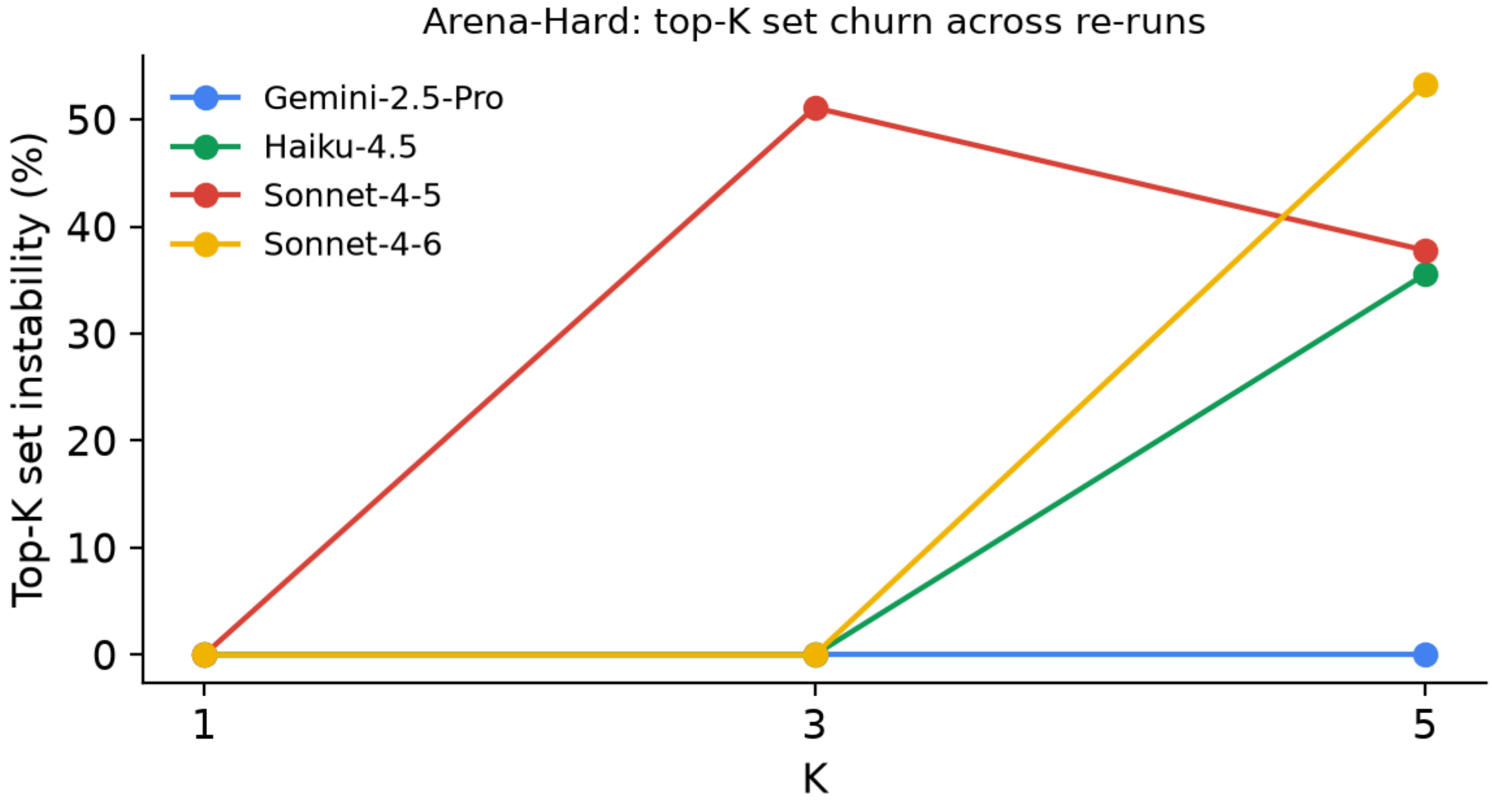

**Figure 2.** Top-K set instability across identical re-runs on Arena-Hard (§6.5c). Gemini is flat at zero for every K; Haiku is zero at top-1 and top-3 but its top-5 rises to 36% as plotted (36–56% depending on how exact ties at the 5/6 boundary are broken, §6.5c); Sonnet-4-5 reshuffles its top-3 (51% as plotted; 51–62% over tie-breaks) and top-5 (38%) sets, and Sonnet-4-6 its top-5 (53%). Plotted values are the lower ends of the tie-break ranges. Aggregate winner-flip stays 0% for all judges.

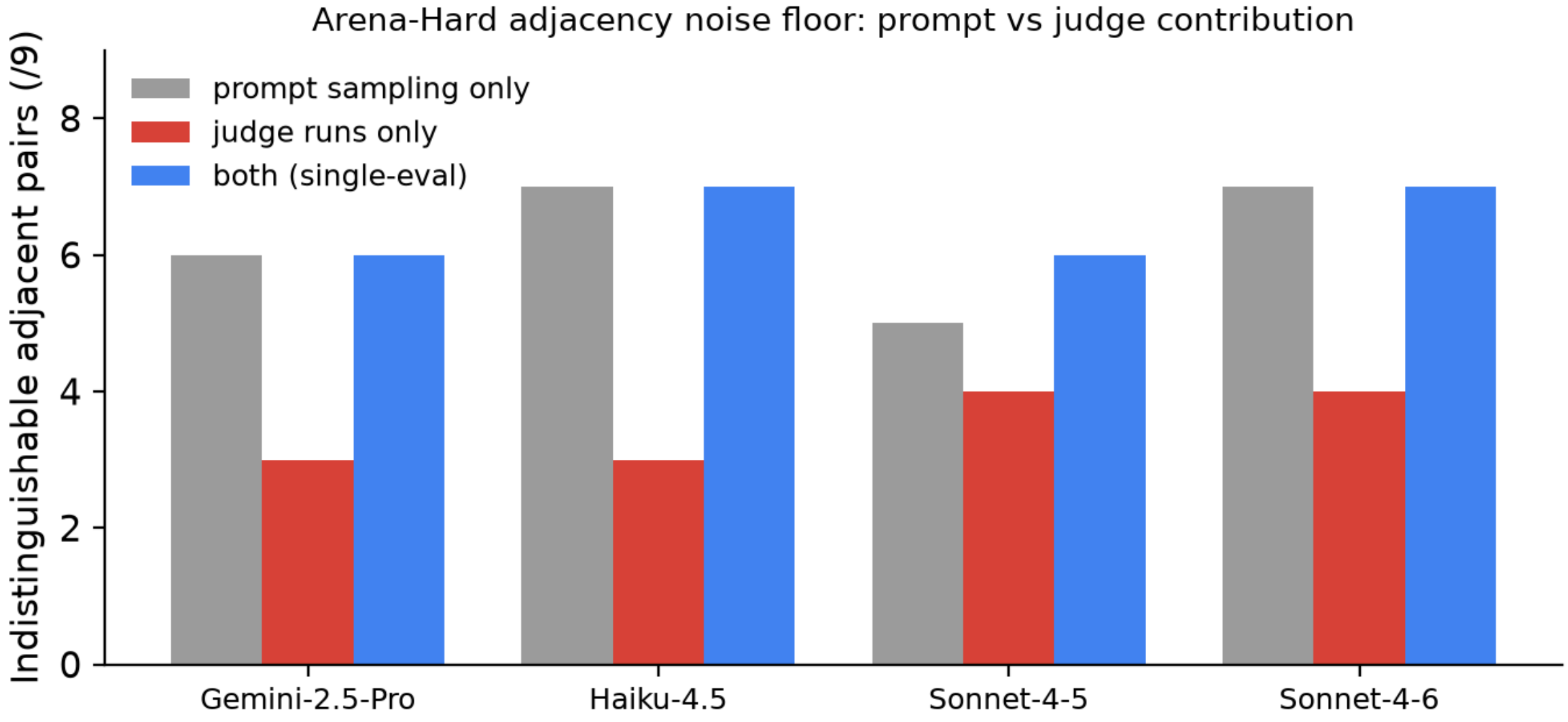

**Figure 3.** Arena-Hard adjacency noise floor decomposed (§6.4). Prompt sampling alone (grey) already produces almost the entire floor; single-run judge variance alone (red) blurs 3–4 of 9 pairs but adds at most one pair *beyond* prompt sampling (blue ≈ grey), so the combined single-eval floor is prompt-dominated.

The pairwise outcome-flip heatmap is omitted as degenerate — aggregate outcome flip is 0% for every judge — and the cross-judge Kendall's $\tau$ values are given in the §6.5d table rather than as a separate figure. Table 1 lists replication survival across judges. (Figures produced by `figures.py`.)

## 7. Discussion

### 7.1 The Fine-Grained Ordering Is the Least Reliable Part

The instability we document is not uniform across the leaderboard; it concentrates where competing models are most similar in true quality. The full sweep (§6) makes this concrete. The top of each leaderboard is comparatively robust — the #1 model is identical across all re-runs of every judge — but the fine-grained ordering below it is not: the conditional flip intensity on close-call items is ~40% for every judge (many times the population mean — roughly 8× for the noisier judges, and over 100× for near-deterministic Haiku, whose 34–35% conditional intensity sits atop a 0.13–0.24% per-item mean — Arena-Hard 0.13%/34.0%, AlpacaEval 0.24%/35.1%), 2–7 of 9 adjacent positions are statistically indistinguishable (§6.4), the noisier judges reshuffle their own top-3/top-5 sets across identical re-runs (§6.5c), and Gemini and the Sonnet judges reorder ≈18% of Arena-Hard model pairs (Gemini vs. Sonnet-4-5, significant) — up to ≈29% for the Gemini/Sonnet-4-6 pair, though that estimate is not statistically significant (§6.5d) — all concentrated in the same close-call region. (A decomposition attributes the adjacency noise floor *primarily* to prompt sampling — a deterministic judge would show nearly the same 2–7 of 9 indistinguishable pairs, with single-run judge variance adding at most one further pair, §6.4 — so this particular effect is largely a benchmark property; the judge-specific instability shows up in item flips, cross-judge disagreement, and replication.) Yet that region is where consumers — model selectors, paper reviewers, marketing teams — tend to read the most into small rank gaps, so an unhedged fine-grained ranking is least reliable where it matters most.

### 7.2 Why Pinning and T=0 Are Not Enough

Three mechanisms documented in prior work plausibly explain the determinism failure; we do not isolate them here — our black-box endpoints only let us measure the *effect* — but each is known to make temperature-zero cloud inference non-reproducible: (1) GPU floating-point non-associativity under variable batch sizes, (2) mixture-of-experts routing dependent on co-located requests (first articulated for production GPT-4 by Chann, 2023), and (3) provider-side load balancing across hardware revisions. None is visible to the API consumer; all would inject variance that propagates through judge logits to discrete win/loss outcomes. We therefore treat these as *hypothesized* sources of the variance we measure, not as causes we establish.

### 7.3 A Minimal Protocol

We propose:

1. **R≥5 judge re-runs** for any published leaderboard entry. Cost is linear in $R$ and negligible relative to candidate-model *training* costs — though we note the trade-off is less favorable for evaluation-heavy uses (prompt engineering, hyperparameter search, small parameter-efficient fine-tunes) where judging is the dominant compute expense; there $R$ re-runs are a real $R\times$ multiplier on evaluation cost, and $R = 5$ rather than $R = 10$ is the pragmatic floor. We note that $R = 5$ leaves per-item rates poorly constrained — a $0/5$ win-count gives a 95% Wilson interval on that item's win *probability* of roughly $0$–$45\%$, which in turn leaves its flip rate weakly determined — so $R = 5$ is the floor for credibility; $R = 10$ tightens these intervals and is what we use in this paper. §8 discusses why even $R = 10$ remains insufficient for distinguishing very small effects. Pin and report the full decoding configuration alongside $R$ — snapshot, temperature, and, for reasoning-capable judges, the exact reasoning/thinking token budget, which affects stability (§8).
2. **Report a stability profile, not just a single flip-rate scalar.** Report (a) the **waver fraction** — the share of items where the judge ever changed verdict across re-runs — and (b) the **conditional intensity** — the mean flip rate restricted to wavering items — separately. Because conditional intensity is near-constant across judges (~34–40%, an $R$-dependent reference set by the close-call ambiguity distribution rather than a judge property; §6.5a, §7.5), the mean flip rate does *track* the ~40× spread rather than hiding it; the value of the two-part profile is therefore *interpretive* — it shows that judges differ in *how often* they waver, not in *how decisively*, and it stops readers from over-reading the conditional intensity as a judge-specific quantity.
3. **Publish the full per-run score matrix**, not just aggregates, so downstream consumers can compute their own intervals using their preferred bootstrap procedure.
4. **Treat two adjacent leaderboard models as tied when the paired bootstrap cannot separate them at the chosen confidence** — on the *paired* score difference (§3.4), i.e. order-preservation below 0.975 for a two-sided 95% test (below 0.95 for one-sided 95% / two-sided 90%). This is the correct separability test, and it is worth distinguishing from two common shortcuts that fail in *opposite* directions. Comparing the two models' *marginal* confidence intervals ignores that they are scored on the same prompts against the same fixed baseline — their scores are positively correlated — and so **overstates** the number of ties. Conversely, asking whether each model's win-rate-*vs-baseline* interval overlaps 50% is not a candidate-vs-candidate test at all and **understates** ties between candidates: a model at 80% and its neighbor at 78% against the baseline both sit well above 50% (so neither is "tied with the baseline"), yet they may be statistically indistinguishable *from each other*. Only the paired test on the two candidates' score difference answers the right question. For a leaderboard that follows point 1 and reports an $R$-run-averaged score, apply this separability test to that averaged metric (resampling the $R$ collected runs, as in point 5); the single-run inner draw of §3.4 is specifically for *diagnosing* existing one-run-per-prompt leaderboards. (We also caution against the rule "tie if the observed verdict-flip rate exceeds 50%": for a binary win/loss the population flip rate $2p(1-p)$ peaks at $0.5$; even with a ternary A/B/tie outcome the maximum probability that two independent draws differ is $1-\sum_k p_k^2 = 2/3$, attained when the three outcomes are equally likely — so an observed rate near either ceiling reflects maximum ambiguity — the judge is at or near maximum entropy over the outcomes — not a decisive signal, and not a small-sample artifact.)

5. **For comparison papers:** report the *probability* that the proposed method outscores the specific model it is being compared against (its competitor, not the fixed reference baseline), computed via the paired hierarchical bootstrap over prompts (§3.4), **resampling the $R$ judge runs you collected within each prompt so the interval matches the $R$-run-averaged score you report** — not a single run. (The single-run inner bootstrap of §6.4 serves a different purpose: it *diagnoses* the noise floor of the standard one-run-per-prompt leaderboards already in the literature. Once you follow point 1 and collect $R \geq 5$ runs, your reported score is the $R$-run mean, and the bootstrap must resample those $R$ runs to reflect its uncertainty. In practice the $m$-vs-$m'$ comparison can be operationalized either as the baseline-anchored win-rate difference $\Delta_i = W_i(m, \text{anchor}) - W_i(m', \text{anchor})$ — what our replication uses (§6.5), needing no extra judge calls — or as fresh head-to-head $m$-vs-$m'$ verdicts; we recommend the former for consistency with how leaderboards already rank.)
6. **Defeat provider-side caching.** Ensure re-runs are not silently served from a cache, which would return identical verdicts and deflate the measured variance. Prefer API-level cache-disabling flags or headers where the provider exposes them; a prepended semantically-neutral nonce is a fallback, but note it perturbs the input prompt and so folds a small amount of prompt-sensitivity into the measurement rather than isolating pure system-level non-determinism (§3.6).
7. **Do not rely on a single judge.** Because judges can disagree on ≈18–29% of Arena-Hard pairwise orderings (§6.5d, Gemini vs. Sonnet) and one-third of the orderings a single judge reproduces fail under a judge from a second family (§6.5), report results under at least two judges from different families — or explicitly scope the claim to the named judge and disclose that a judge swap may change it.
8. **For scalar benchmarks (e.g. MT-Bench):** map the protocol to absolute scores — report per-run mean scores with their bootstrap interval, define a "waver" as a change in the leadership sign between two models across re-runs (§3.4), and treat two models as tied when the bootstrap interval of their mean-score difference includes zero. The R≥5 and stability-profile recommendations apply unchanged.

### 7.4 Implications for Existing Literature

We do not claim that the body of published model comparisons is wrong — and our replication set does not test it directly: it consists of 15 expected orderings (vendor tiers, model generations, and leaderboard standings), not results reported on these benchmarks (§3.5). Of these, **8 survive every judge that evaluated them and all re-runs** — evidence that the larger capability gaps are robust to *judge* variance. (This robustness is distinct from clearing the *prompt-sampling* noise floor of §6.4: the replication test re-judges a fixed prompt set, so a gap can survive all judges yet still be statistically indistinguishable under prompt resampling. The two are complementary hedges, and a claim needs to clear both.) We do claim that a substantial minority do not: **7 of the 15 fail** — 3 because judges disagree on the winner, 2 because the same judge flips across identical re-runs (including the Llama-4-Maverick > Llama-3.3-70B generation ordering, which Gemini reverses on 4 of 10 re-runs), and 2 (#3, #12) because every judge ranks the other model higher — one (#3) robust to length control, the other (#12) length-confounded (§6.5). For comparisons reporting sub-3-percentage-point Arena-Hard win-rate gaps or sub-0.2-point MT-Bench differences — inside the noise floor of §6.4 — the published claim is not supported (these thresholds are empirical, set by the benchmark prompt counts — 500 for Arena-Hard, 80 for MT-Bench — and this candidate pool's capability density, not universal constants; §6.4, §8). We are careful about attribution: that noise floor is driven primarily by the *finite prompt sample* (a gap this small is statistically fragile even under a perfectly deterministic judge, §6.4), with judge run-to-run variance and judge *choice* adding further instability on top. Both sources must be hedged — reporting judge re-runs alone would not fix a gap that prompt sampling already cannot resolve. Finally, the within-judge variance estimates here are *lower bounds*: we hold candidate outputs fixed (§3.6), whereas real evaluation pipelines regenerate them, adding candidate-generation variance on top of the judge-side variance we measure. (This does not extend to the Arena-Hard cross-judge comparisons, which truncated candidate answers can move in either direction; §8.) A published ordering already fragile under these controlled, variance-suppressed conditions can only be more fragile in end-to-end practice.

### 7.5 Why Judge Stability Is a Profile, Not a Number

The full sweep confirms and sharpens the pilot finding that judges differ in *how* they are unstable. Per-item flip rate spans a ~40× range across judges served on the same infrastructure — `claude-haiku-4-5` flips only 0.13–0.24% of verdicts while `claude-sonnet-4-5` flips 5–10% (§6.5a) — while the *conditional* intensity is near-uniform (34–40%). As §6.5a shows, that near-uniformity reflects a close-call ambiguity distribution that is broadly spread (only ~26% of wavering items are single-run wavers) and similar across judges — its $R$-dependent reference value is $(R+1)/(3(R-1)) \approx 40.7\%$ at $R = 10$ — so the entire ~40× spread is carried by the **waver fraction**, not the conditional intensity.

Two consequences for reporting follow. First — correcting an earlier framing — because the conditional intensity is near-constant, the mean per-item flip rate is approximately proportional to the waver fraction, so a single mean-flip-rate scalar does *not* hide the ~40× spread; it tracks it. The value of reporting the two components separately is therefore *interpretive* rather than information-preserving: it makes explicit that judges differ in *how often* they waver, not in *how decisively*, and prevents misreading a benchmark's headline flip rate. Second, both components are $R$-dependent — the waver fraction can only grow with $R$, and the conditional-intensity reference value scales as $(R+1)/(3(R-1))$ (§6.5a) — so both must be reported at a stated $R$ to be comparable across studies. A caveat for near-deterministic judges: Haiku's Arena-Hard conditional intensity is estimated from the ~0.4% of 5,000 cells (≈19) that ever wavered, so its precise value is indicative, not exact.

## 8. Limitations

- We measure judge variance with candidate outputs held fixed. Real workflows regenerate candidates, adding further variance not captured here.
- **Some candidate answers reached the generation cap.** Candidates were generated with `max_output_tokens=2048` (§3.3), and some of the longer Arena-Hard answers (by a conservative count, about 13%) were cut off before completion; AlpacaEval and MT-Bench are barely affected. Judges may score incomplete answers differently, so the Arena-Hard cross-judge comparisons (§6.5d), including the lowest cross-judge $\tau$ and the #9 disagreement, are sensitive to this and should be read with that caveat. The within-judge results (§6.2–6.4, §6.5a–c) compare verdicts on identical inputs across re-runs and are unaffected in kind, though their magnitudes are specific to this candidate set. Regenerating the affected answers with a larger cap and re-judging them is the natural follow-up.
- **No cache-buster was used, which could confound the cross-judge magnitudes.** We did not disable provider-side caching (§3.6), so all flip rates are lower bounds. This matters most for the cross-*judge* comparison: if cache-hit rates happen to differ across endpoints (which we neither controlled nor measured), part of the ~40× stability spread (e.g. Haiku's near-determinism) could reflect caching rather than intrinsically steadier judging. Cache-disabling flags (or a nonce, with the caveat in §7.3) would re-move this confound; the *existence* of nonzero within-judge variance is unaffected, since a full cache would drive it to exactly zero.

- Our R=10 judge re-runs are sufficient to demonstrate flip rates but not to establish tight confidence intervals on small effects.
- We do not measure human-judge agreement; the question we ask is internal consistency of automated judges, not their correctness.
- Provider-side determinism behavior may change after publication; we report on the snapshots and access patterns valid at the time of the experiment.
- **Judges were invoked by alias, not by dated snapshot identifier.** The serving-reported version string was constant within each judge (§3.2), but a provider can change serving internals without changing that string. This is the same position as any practitioner who pins a judge version, which is the setting this paper studies.
- The replication set is small (15 orderings among models in our pool) and was chosen by us rather than sampled from published comparisons, so its survival rate should not be read as a population-wide replication rate.
- **All judges are served via a single enterprise cloud platform (§3.2).** We do not directly measure within-snapshot variance of GPT-4-class judges hosted by OpenAI. We expect the qualitative result to transfer because the variance mechanisms (floating-point non-associativity in batched matmul, MoE routing dependent on co-located requests, hardware-revision drift) are not provider-specific, and prior work has reported T=0 non-determinism across multiple LLM families (Atil et al., 2024) and analogous within-judge inconsistency on GPT-class models (Stureborg et al., 2024; Wang et al., 2023). We flag confirmation on GPT judges as straightforward future work. This single-platform design also confounds the *cross-family relative* stability gap: that Claude Sonnet judges are more unstable than Gemini (§6.5a) could in principle include a serving-stack contribution rather than being purely an intrinsic property of the Claude models; with black-box endpoints we cannot separate the two, and we do not attribute it to the platform. The *existence* of per-judge variance is robust; its precise cross-family ordering should be read as "as served on this platform," and confirming it against each vendor's native API is the natural next step.
- The replication claims are expected orderings — from vendor model tiers, model generations, public leaderboard standings, and, for #3 and #12, our own expectation (Appendix C) — not results reported on these benchmarks. We re-judge each under our four-judge ensemble (all served via the same single platform) and separate the failure mechanisms (§6.5, Table 1) into within-judge instability, cross-judge disagreement, and unanimous contradiction. A failure means the expected ordering is not robust to reruns and a reasonable judge swap under proper variance accounting — not that any reported number is wrong. Under length control (§6.5) #3's contradiction is robust (the preferred model wins even length-matched) while #12's is length-confounded (Claude Opus 4.5's outputs are ~43% shorter than Qwen3-235B's), so only one is a clean benchmark-level contradiction.
- Our candidate pool is 10 models across four families (Gemini, Claude, Llama, Qwen). The full Stage 2 sweep (§6) confirms the effect well beyond the pilots' Flash-vs-Flash-Lite pair — across all 10 candidates and all four judges — but generalization to arbitrary models outside this pool remains to be shown.
- `claude-sonnet-4-6`'s AlpacaEval cell was cut short when we stopped the sweep to bound *total* spend (completing that one cell alone would be inexpensive, §6.5); that judge's AlpacaEval results are excluded, so its half of the matched-snapshot comparison rests on Arena-Hard and MT-Bench only.
- **All evaluations are single-turn, and context-free turn-2 items inflate MT-Bench variance.** MT-Bench's two-turn prompts are split into independent single-turn items, and turn-2 items are judged without turn-1 context (§3.1). Empirically these turn-2 items are about **twice as unstable** as turn-1 items across every judge (§6.2), consistent with their being underspecified without context — so the pooled MT-Bench figure is partly an artifact of this design choice, and the turn-1-only rate is the cleaner (lower) estimate. Multi-turn judging, where an unstable turn-1 verdict propagates into turn-2, may also have a different profile; our results characterize single-turn judgments only.

- **Position bias is ruled out only on a disparate-tier pair.** The §4.2 position-swap check used Flash vs. Flash-Lite, where a consistent quality preference could mask a residual position effect; prior work notes position bias is strongest for near-tie pairs. We did not run a full-scale position swap on the near-tie pairs where conditional flip rates are highest, so a small position component there cannot be fully excluded. Moreover, every Stage 2 comparison used the same fixed order (reference answer first, candidate second; §3.4), so a position preference, if present, would not average out and could shift absolute win-rates. A *targeted* swap on a handful of the near-tie pairs (rather than a full-benchmark swap, which would roughly double judging cost) would settle this cheaply and is the natural next check.
- **The principal judge ran with a minimal reasoning budget.** `gemini-2.5-pro` was configured with `thinking_budget=128` (the platform minimum) throughout both the pilots and the Stage 2 sweep. We chose the minimum for cost and throughput at the sweep's scale, not for a principled reason, and we did not sweep the budget — a limitation, since a larger budget could change its stability profile in *either* direction — *suppressing* variance if the minimal budget was truncating a stabilizing chain of thought (so restoring it would stabilize the judge, and would plausibly also remove the degenerate-generation pathology of §4.3), or *inflating* variance if a longer budget widens the space of rationales the judge samples — and we cannot determine the sign from black-box endpoints without a default-budget re-run. That re-run is a cheap, targeted follow-up (a 50-prompt × 10-run probe would settle the direction), which we flag as future work. Gemini's *absolute* flip rates should therefore be read as specific to this configuration; the Claude judges have no comparable setting, and the paper's cross-judge results — the ~40× stability spread, the cross-judge $\tau$ disagreements, and the replication overturns — do not depend on Gemini's absolute rate. (Mechanistically the two directions are not symmetric a priori: a longer reasoning trajectory both avoids truncation pathologies and gives token-level numerical non-determinism, §7.2, more steps over which to compound before the verdict field.)
- **The adjacency noise floor is primarily a prompt-sampling effect, not judge instability.** A prompt-only bootstrap reproduces almost the entire noise floor, and adding single-run judge-run resampling contributes at most one further indistinguishable pair (+1 in one of seven judge×benchmark cells; §6.4), so we attribute it primarily to benchmark prompt sampling. This is a scoping clarification rather than an open limitation: the judge-specific results (§6.2, §6.5) stand on their own, and the noise floor bounds leaderboard precision regardless of its source.

## 9. Conclusion

LLM-as-Judge evaluation commonly treats a pinned, temperature-zero judge as a noiseless instrument. Our results show that it is not. Across three benchmarks and four frontier judges from two model families (Gemini and Claude), all served via a single enterprise cloud platform (§3.2), pinned temperature-zero judges produce per-item verdict variance whose magnitude is a per-judge property — as served on this platform — spanning ~40× (0.13% for Haiku to 5–10% for Sonnet-4-5) and ~40% on close-call items, and not explained by a consistent position preference (§4.2; a residual position component on the closest pairs is not fully excluded, §8). This variance does not invert the reference judge's aggregate ranking, but it leaves between a fifth and three-quarters of adjacent positions statistically indistinguishable (a noise floor driven *primarily by the finite prompt sample*, with single-run judge variance adding at most one further pair; §6.4), reshuffles the noisier judges' own top-K sets across identical re-runs, and — across judges — re-orders a significant ≈18% of Arena-Hard model pairs between Gemini and Sonnet-4-5 (up to ≈29% for the Gemini/Sonnet-4-6 pair, though that estimate is not statistically significant). Of 15 expected head-to-head orderings, 12 survive every re-run of the principal judge but only 8 survive every judge — a second judge overturns one-third of what a single judge reproduces — and 2 are contradicted by every judge. The fix is straightforward and cheap: re-run the judge a small number of times, report the stability profile (waver fraction and conditional intensity, §7.5) and adjacency separability, judge under more than one family, and report point estimates together with their uncertainty. Until this becomes standard, leaderboards should be read as approximate orderings with implicit ties of unknown width — and with a footnote naming the judge.

## Appendix A — Reproducibility

The pilot studies (§4) cost approximately **$33** in platform API spend and ran in under three hours wall-clock on a single workstation with Application Default Credentials. The full Stage 2 sweep covered four judges × 10 candidates × 3 benchmarks at R=10 (with `claude-sonnet-4-6` completing Arena-Hard and MT-Bench only), for platform spend on the order of **$5,000** (a position-swap pass, not run at scale, would roughly double pairwise judging cost). It ran on a single VM; the binding constraint was our project's default request-rate quota on the platform's Anthropic endpoints (~2–3 requests/second at that quota — a per-project limit that can be raised on request, not a platform throughput ceiling), giving roughly 18 hours of wall-clock per judge.

The runner code, the schema-enforced verdict-extraction wrappers (`response_schema` for Gemini judges, forced `tool_use` for Anthropic judges), the per-run verdict data, the candidate response cache, and the prompt templates are available on reasonable request, and a cleaned public release is in preparation — the per-run verdict data and the analysis code (which reproduce every number and figure directly from the raw verdicts) as the priority artifact, packaged as a GitHub repository for the code and a HuggingFace dataset for the verdict trajectories and candidate responses. Links will be added in a later revision of this paper once the release clears internal review.

## Appendix B — Per-Judge Identification (for Pilot Studies §4)

For consistency with the family-neutral framing in §4.2, the per-judge results table anonymized the two judges as $J_1$ and $J_2$. The mapping is:

- $J_1$: `gemini-2.5-pro` (the platform, region `us-central1`, `temperature=0`, `thinking_budget=128`, `max_output_tokens=4096`)
- $J_2$: `claude-haiku-4-5@20251001` (the platform, global endpoint via its Anthropic offering, `temperature=0`, `max_output_tokens=4096`)

Both judges share identical prompt templates, candidate inputs, and evaluation pipelines; the only difference at the judge stage is model identity. Verdict extraction in the pilot studies used **free-form text parsing** for both judges (`[[A]] | [[B]] | [[C]]` for Tier 0, `A | B | tie` for Tier 1) — structured-output enforcement (`response_schema` for Gemini, forced `tool_use` for Anthropic) was *not* used in the pilots. The 24% parse-failure rate on $J_1$ Tier 1 documented in §4.3 is a direct consequence of this text-format choice; structured-output enforcement is adopted starting with the Stage 2 sweep (§5). Pilot 1 (§4.1) used only $J_1$.

## Appendix C — Per-Claim Replication Detail (for §6.5)

Fulfilling the §5.4 reporting protocol, we give per-claim, per-judge outcomes for all 15 replication claims. Each cell is **first-run reproduced (Y/N) / fraction of the 10 re-runs that reproduced the claim**; "—" marks the excluded `claude-sonnet-4-6` × AlpacaEval cell. A claim *survives* a judge only when the fraction is 1.0.

| # | Claim | Benchmark | Gemini | Haiku | Sonnet-4-5 | Sonnet-4-6 |
|---|---|---|---|---|---|---|
| 1 | Gemini 2.5 Pro > Flash-Lite | Arena-Hard | Y/1.0 | Y/1.0 | Y/1.0 | Y/1.0 |
| 2 | Claude Opus 4.5 > Haiku 4.5 | Arena-Hard | Y/1.0 | Y/1.0 | Y/0.9 | Y/1.0 |
| 3 | Qwen3-235B > Qwen3-Next-80B | AlpacaEval | N/0.0 | N/0.0 | N/0.0 | — |
| 4 | Gemini 2.5 Pro > Flash | Arena-Hard | Y/1.0 | Y/1.0 | Y/1.0 | Y/1.0 |
| 5 | Gemini 2.5 Flash > Flash-Lite | AlpacaEval | Y/1.0 | N/0.0 | Y/0.9 | — |
| 6 | Claude Opus 4.5 > Opus 4.1 | Arena-Hard | Y/1.0 | Y/0.5 | N/0.4 | N/0.4 |
| 7 | Claude Opus 4.1 > Haiku 4.5 | AlpacaEval | Y/1.0 | Y/1.0 | Y/1.0 | — |
| 8 | Llama-4-Maverick > Llama-3.3-70B | AlpacaEval | Y/0.6 | Y/1.0 | Y/1.0 | — |
| 9 | Gemini 2.5 Pro > Qwen3-235B | Arena-Hard | Y/1.0 | N/0.0 | N/0.0 | N/0.0 |

| # | Claim | Benchmark | Gemini | Haiku | Sonnet-4-5 | Sonnet-4-6 |
|---|---|---|---|---|---|---|
| 10 | Claude Opus 4.5 > Llama-4-Maverick | Arena-Hard | Y/1.0 | Y/1.0 | Y/1.0 | Y/1.0 |
| 11 | Gemini 2.5 Pro > Llama-4-Maverick | AlpacaEval | Y/1.0 | Y/1.0 | Y/1.0 | — |
| 12 | Claude Opus 4.5 > Qwen3-235B | AlpacaEval | N/0.0 | N/0.0 | N/0.0 | — |
| 13 | Qwen3-235B > Llama-3.3-70B | Arena-Hard | Y/1.0 | Y/1.0 | Y/1.0 | Y/1.0 |
| 14 | Gemini 2.5 Pro > Flash-Lite | MT-Bench | Y/1.0 | Y/1.0 | Y/1.0 | Y/1.0 |
| 15 | Claude Opus 4.5 > Haiku 4.5 | MT-Bench | Y/1.0 | Y/1.0 | Y/1.0 | Y/1.0 |

Reading the failure mechanisms of §6.5 off this table — where, per (claim, judge), the *aggregate* winner is the claimed model when the fraction exceeds 0.5, the other model when it is below 0.5, and undetermined at exactly 0.5: **within-judge instability** is a reproduced fraction strictly between 0 and 1 with the aggregate winner still the claimed model (Claim 8 under Gemini, 0.6; Claim 2 under Sonnet-4-5, 0.9); **cross-judge disagreement** is judges reaching *different aggregate winners* — at least one with fraction > 0.5 and at least one with fraction ≤ 0.5 — for Claims 5, 6, 9 (Claim 6 under Haiku sits at exactly 0.5, a judge with no stable aggregate winner, which we place on the non-reproducing side); **unanimous contradiction** is a reproduced fraction of 0.0 across all judges with data (Claims 3, 12). A note on classification vs. observation. The three labels above classify a *claim* by the pattern across judges; they are distinct from the generic observation that an individual judge *wavers* (any fraction strictly in $(0, 1)$). Claims 5 and 6 are classified as **cross-judge disagreement** because judges reach different aggregate winners, and on some judges they *also* show judge-level wavering — but note this wavering does not always meet the formal **within-judge-instability** definition, which additionally requires the aggregate winner to remain the claimed model (fraction $> 0.5$). Claim 6 illustrates the distinction: Gemini 1.0, Haiku 0.5, Sonnet-4-5 0.4, Sonnet-4-6 0.4 — no single judge here has a fraction strictly in $(0.5, 1)$, so Claim 6 is a pure cross-judge disagreement, not a within-judge-instability case, even though individual judges waver. We classify each claim by its dominant pattern in Table 1. The basis for each claim's expected ordering is given in the provenance table below. The two MT-Bench claims (#14, #15) are scored on the mean over both turns, which includes turn-2 items judged without their turn-1 context (§3.1). Produced by `replication_from_sweep.py` / `noise_floor_decomposition.py`; the AlpacaEval length-control check (§6.5) by `alpaca_length_control.py`.

**Claim provenance.** The basis for each claim's expected ordering (all orderings evaluated on the models in our candidate pool, §3.3). *Vendor tiering* and *model generation* rows reflect how the vendor positions its own models, not results reported on these benchmarks.

| # | Claim | Basis for the expected ordering |
|---|---|---|
| 1 | Gemini 2.5 Pro > Flash-Lite (Arena-Hard) | Vendor tiering: Gemini 2.5 family (Comanici et al., 2025) |
| 2 | Claude Opus 4.5 > Haiku 4.5 (Arena-Hard) | Vendor tiering: Claude family (Anthropic, 2025) |
| 3 | Qwen3-235B > Qwen3-Next-80B (AlpacaEval) | Our expectation: model size within the Qwen3 family |
| 4 | Gemini 2.5 Pro > Flash (Arena-Hard) | Vendor tiering: Gemini 2.5 family (Comanici et al., 2025) |
| 5 | Gemini 2.5 Flash > Flash-Lite (AlpacaEval) | Vendor tiering: Gemini 2.5 family (Comanici et al., 2025) |
| 6 | Claude Opus 4.5 > Opus 4.1 (Arena-Hard) | Model generation: Claude Opus line (Anthropic, 2025) |

| # | Claim | Basis for the expected ordering |
|---|---|---|
| 7 | Claude Opus 4.1 > Haiku 4.5 (AlpacaEval) | Vendor tiering: Claude family (Anthropic, 2025) |
| 8 | Llama-4-Maverick > Llama-3.3-70B (AlpacaEval) | Model generation: Llama 4 (Meta AI, 2025) vs. Llama 3.3 (Meta AI, 2024) |
| 9 | Gemini 2.5 Pro > Qwen3-235B (Arena-Hard) | Public Arena-Hard/LMArena leaderboard standing (Li et al., 2024; Chiang et al., 2024), accessed 2026-08-14 |
| 10 | Claude Opus 4.5 > Llama-4-Maverick (Arena-Hard) | Public Arena-Hard/LMArena leaderboard standing (Li et al., 2024; Chiang et al., 2024), accessed 2026-08-14 |
| 11 | Gemini 2.5 Pro > Llama-4-Maverick (AlpacaEval) | Public AlpacaEval 2 leaderboard standing (Dubois et al., 2024), accessed 2026-08-14 |
| 12 | Claude Opus 4.5 > Qwen3-235B (AlpacaEval) | Our expectation: flagship proprietary model vs. open-weights model |
| 13 | Qwen3-235B > Llama-3.3-70B (Arena-Hard) | Public Arena-Hard/LMArena leaderboard standing (Li et al., 2024; Chiang et al., 2024), accessed 2026-08-14; the Qwen3 technical report (Yang et al., 2025) covers the original Qwen3-235B-A22B, not the Instruct-2507 checkpoint in our pool |
| 14 | Gemini 2.5 Pro > Flash-Lite (MT-Bench) | Vendor tiering: Gemini 2.5 family (Comanici et al., 2025) |
| 15 | Claude Opus 4.5 > Haiku 4.5 (MT-Bench) | Vendor tiering: Claude family (Anthropic, 2025) |

Claims 9–11 and 13 are cross-family orderings we take from public leaderboard standings rather than a primary source; readers should treat them as "leaderboard-plausible at the time of access" rather than authoritative results, and the within-family rows as capability expectations rather than measured results. **Standings reflect the LMSYS / LMArena Chatbot Arena leaderboard snapshot accessed 2026-08-14** (and, for Claim 11, the AlpacaEval 2 leaderboard on the same date); live leaderboards change over time, so these provenance rows are timestamped rather than permanent.

## Appendix D — Changes from v1

This revision corrects and clarifies v1. No new experiments were run, and all within-judge variance results (§6.2–6.4, §6.5a–b) are unchanged.

1. **Framing of the replication orderings (§1, §3.5, §5.4, §6.5, §7.4, Appendix C).** v1 described 13 of the 15 orderings as published claims and 2 as directional expectations we set. v2 describes all 15 as expected orderings and states the basis for each (vendor tiering, model generation, public leaderboard standing, or our own expectation). The counts are unchanged: 12 of 15 survive the principal judge, 8 of 15 survive every judge, and a second judge overturns 4 of the 12 (one-third). Statements that relied on the published/expected split are removed.
2. **Candidate generation cap (§3.3, §8).** v1 stated `max_output_tokens=4096` for candidates; they were generated with 2,048 (4,096 is the judge limit). v2 corrects the value and adds a limitation: some longer Arena-Hard answers reached the cap, which may affect the Arena-Hard cross-judge comparisons. The self-preference reading of #9 is qualified accordingly.
3. **Top-K ties (§3.4, §6.3, §6.5c, Figure 2).** Two top-K values depend on how exact ties at the K boundary are broken and are now reported as ranges (Haiku top-5: 36–56%; Sonnet-4-5 top-3: 51–62%); Figure 2 plots the lower values. The abstract's statement that both Sonnet judges reshuffle their top-3 set on roughly half of re-runs is corrected (Sonnet-4-6's top-3 set is stable), as is §6.3's statement that the top-K sets are stable for all four judges (only the top-1 model is).

4. **Stage 2 data completeness (§4.3, §6.5).** v1 stated 0% parse errors. At least 99.8% of Stage 2 verdicts are valid for every judge × benchmark; the residual gaps are listed in §6.5.
5. **Abort threshold (§4.3).** The per-round parse-error abort threshold is 1% for the first round and 0.5% thereafter, not 5%.
6. **Pinning (§3.2, §3.3, §5.1, §8).** Models were invoked by platform alias rather than a dated snapshot identifier; the serving-reported version was constant within each judge. v2 states this.
7. **Length control for #5 (§3.1, §6.5).** v1 stated that #5's cross-judge disagreement persists under length control. It persists on the length-matched subset but not under the logistic adjustment.
8. **Bootstrap resamples (§3.4).** The adjacency bootstrap behind the reported counts uses 20,000 resamples, not 1,000.
9. **Answer order (§3.4, §8).** Every Stage 2 comparison showed the reference answer first (slot A) and the candidate second (slot B). v1's notation placed the candidate in slot A; v2 corrects the notation and states that the order was fixed rather than randomized. Win-rates were always computed by model identity, so no reported number changes.
10. **Wording.** Claim #12 is described as length-confounded throughout (v1 also called it a length artifact). §2 no longer describes the adjacency noise floor as a judge-variance floor, which contradicted §6.4's attribution to prompt sampling. §7.4's lower-bound statement is limited to the within-judge variance estimates. The §8 limitation on how papers were selected is replaced, since the orderings were chosen by us. Appendix C notes that the MT-Bench claims include turn-2 items. §4.2's cross-family generalization no longer extends to serving infrastructure, since both judges were served on one platform. References to released code and data now point to Appendix A, which describes their availability. Statements that generalized beyond the one platform tested, and an unqualified statement about prior work, are qualified. A few phrasings are made more measured.